\pdfoutput=1
\documentclass[11pt]{article}

\usepackage{ACL2023}

\usepackage{times}
\usepackage{latexsym}
\usepackage{microtype}
\usepackage{graphicx}
\usepackage{subfigure}
\usepackage{booktabs} % for professional tables
\usepackage{hyperref}
\usepackage{amsmath}
\usepackage{amssymb}
\usepackage{float}

\usepackage[T1]{fontenc}
\usepackage[utf8]{inputenc}

\usepackage{microtype}

\usepackage{inconsolata}

\title{Disentangling Length from Quality in Direct Preference Optimization}

\author{Ryan Park* \\
  Stanford University\\
  \texttt{rypark@stanford.edu} \\\And
  Rafael Rafailov*\\
  Stanford University\\
  \texttt{rafailov@stanford.edu} \\\AND
  Stefano Ermon\\
  Stanford University\\
  \texttt{ermon@stanford.edu}  \\\And
  Chelsea Finn\\
  Stanford University\\
 \texttt{cbfinn@stanford.edu} \\}
\begin{document}
\maketitle
\begin{abstract}
Reinforcement Learning from Human Feedback (RLHF) has been a crucial component in the recent success of Large Language Models. However, RLHF is know to exploit biases in human preferences, such as verbosity. A well-formatted and eloquent answer is often more highly rated by users, even when it is less helpful and objective. A number of approaches have been developed to control those biases in the classical RLHF literature, but the problem remains relatively under-explored for Direct Alignment Algorithms such as Direct Preference Optimization (DPO). Unlike classical RLHF, DPO does not train a separate reward model or use reinforcement learning directly, so previous approaches developed to control verbosity cannot be directly applied to this setting. Our work makes several contributions. For the first time, we study the length problem in the DPO setting, showing significant exploitation in DPO and linking it to out-of-distribution bootstrapping. We then develop a principled but simple regularization strategy that prevents length exploitation while still maintaining improvements in model quality. We demonstrate these effects across datasets on summarization and dialogue, where we achieve up to 20\% improvement in win rates when controlling for length, despite the GPT-4 judge's well-known verbosity bias. 
\end{abstract}

\section{Introduction}
{\let\thefootnote\relax\footnote{{*Denotes equal contribution}}}

\begin{figure}
    \centering
    \includegraphics[width=0.525\textwidth]{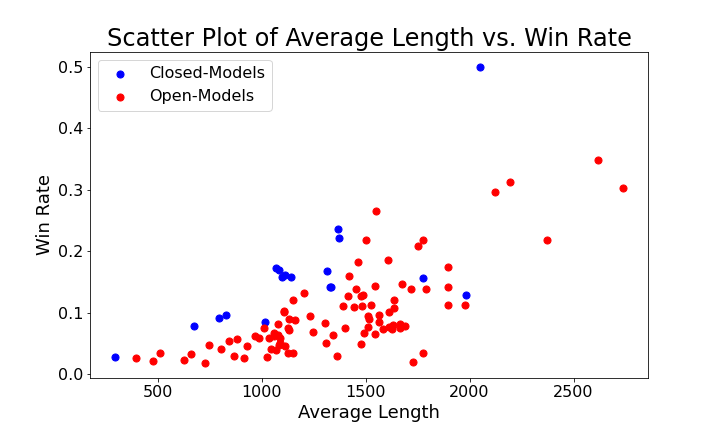}
    \caption{Average win rates versus generation length \cite{tweet2024Liu} on the Alpaca Eval benchmark \cite{dubois2024alpacafarm}. While the highest-scoring open-source models can match the overall performance of strong closed models, they lag significantly on length-corrected basis.}
    \label{fig:length-rank}
\end{figure}

\begin{figure*}
    \centering
    \includegraphics[width=0.225\textwidth]{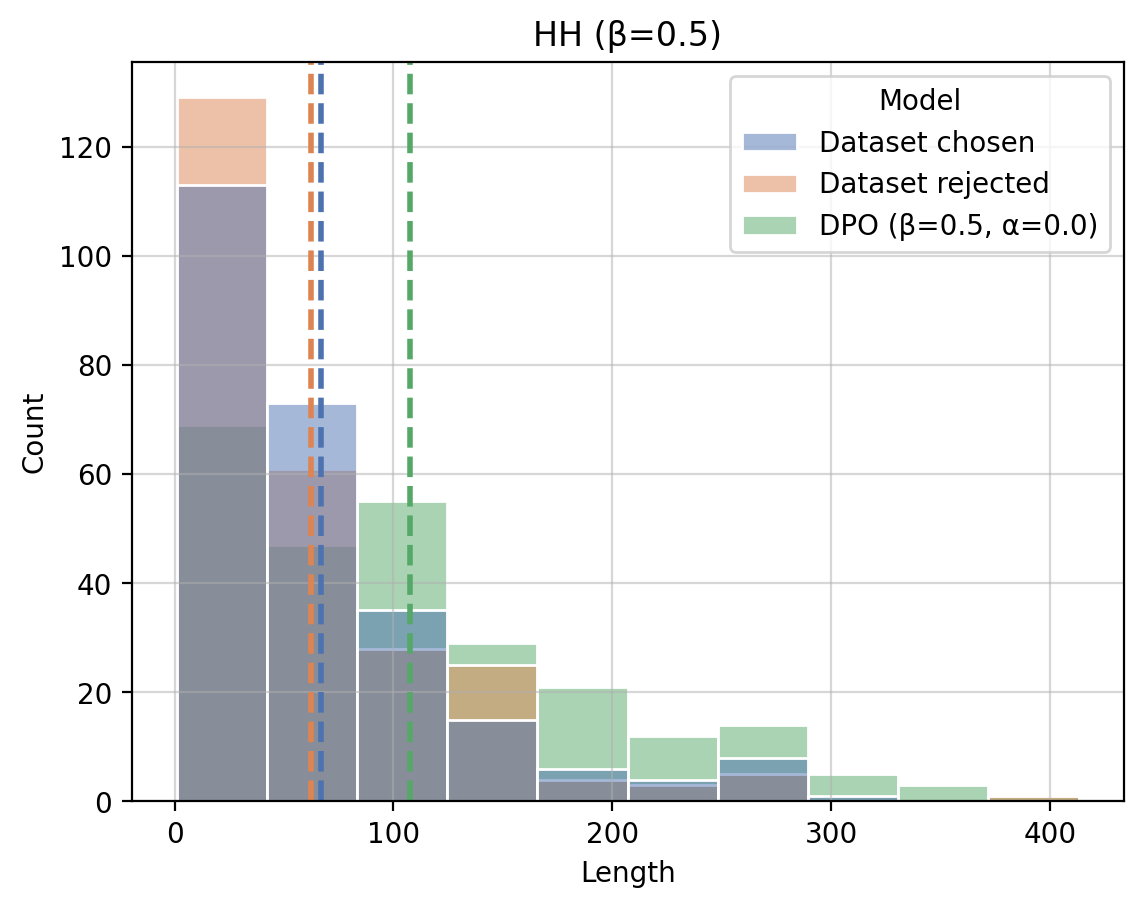}
    \includegraphics[width=0.225\textwidth]{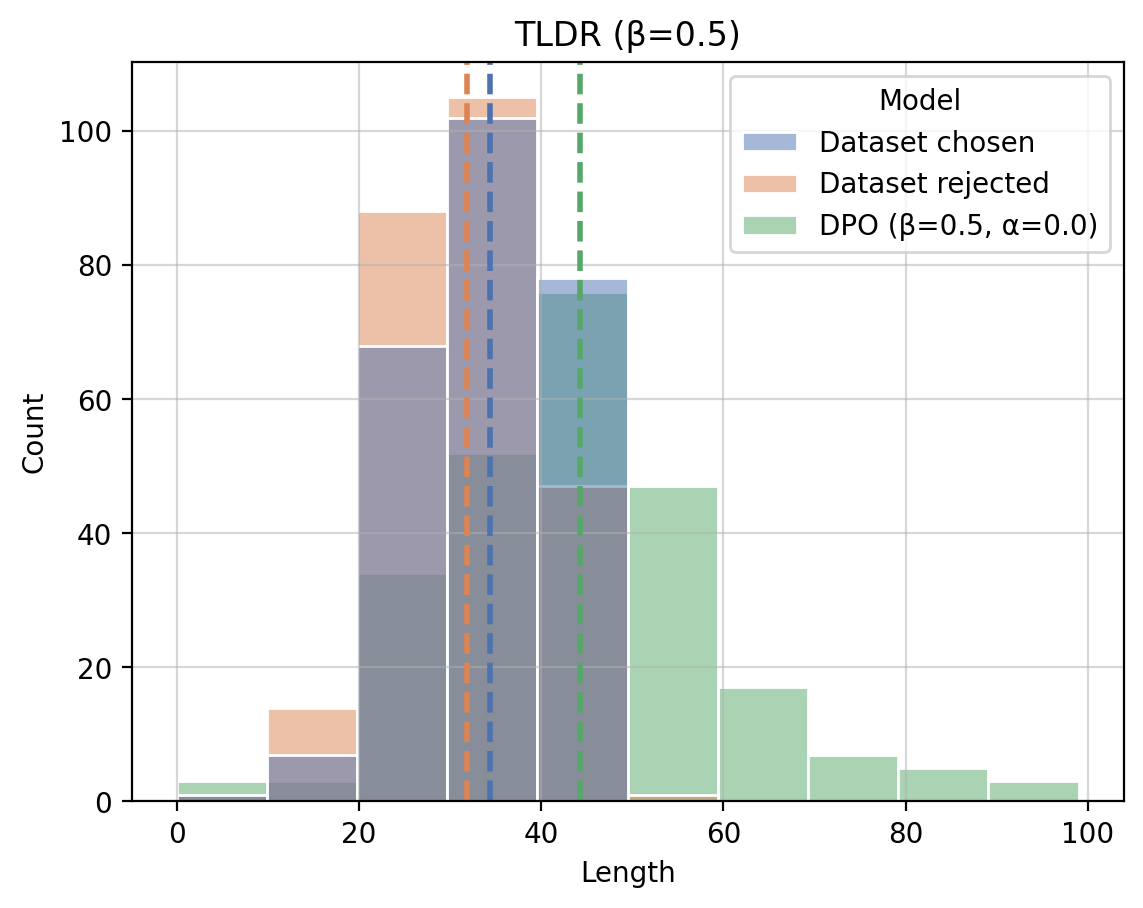}
    \includegraphics[width=0.225\textwidth]{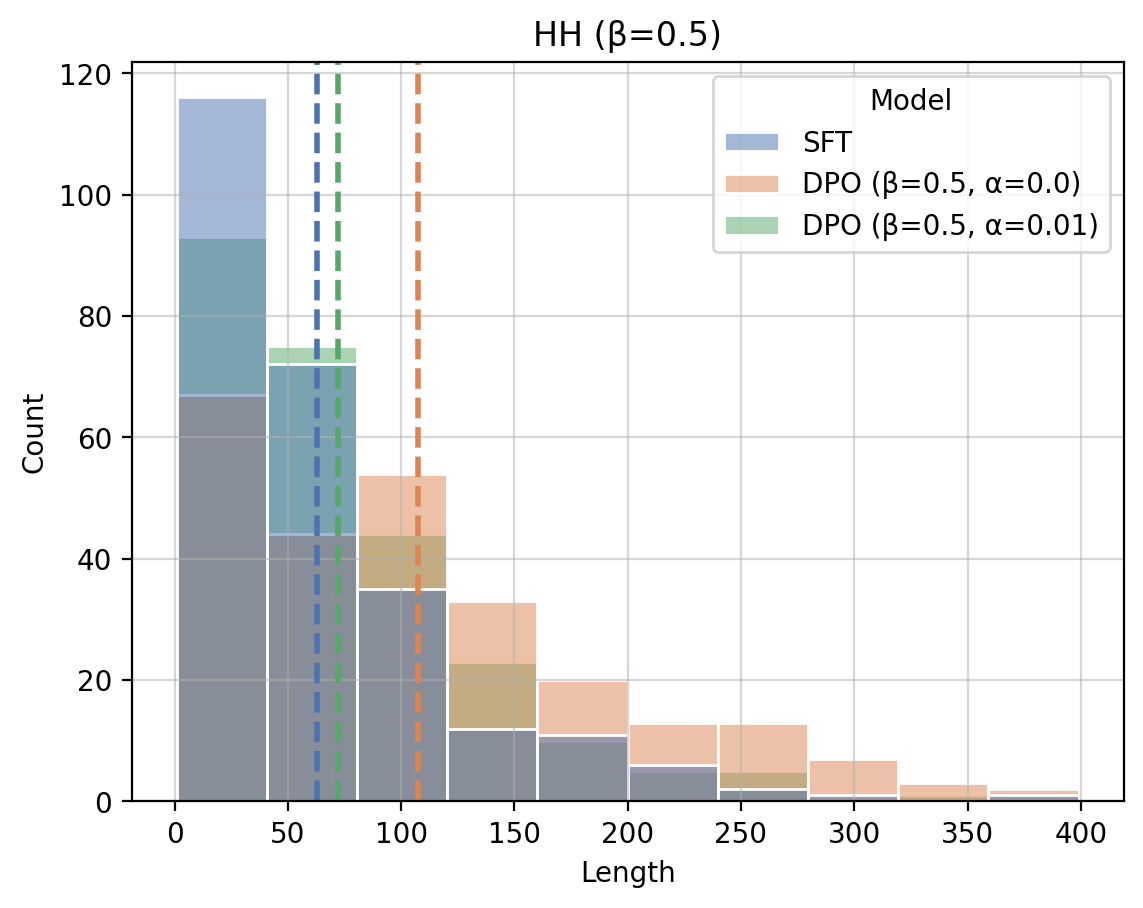}
    \includegraphics[width=0.225\textwidth]{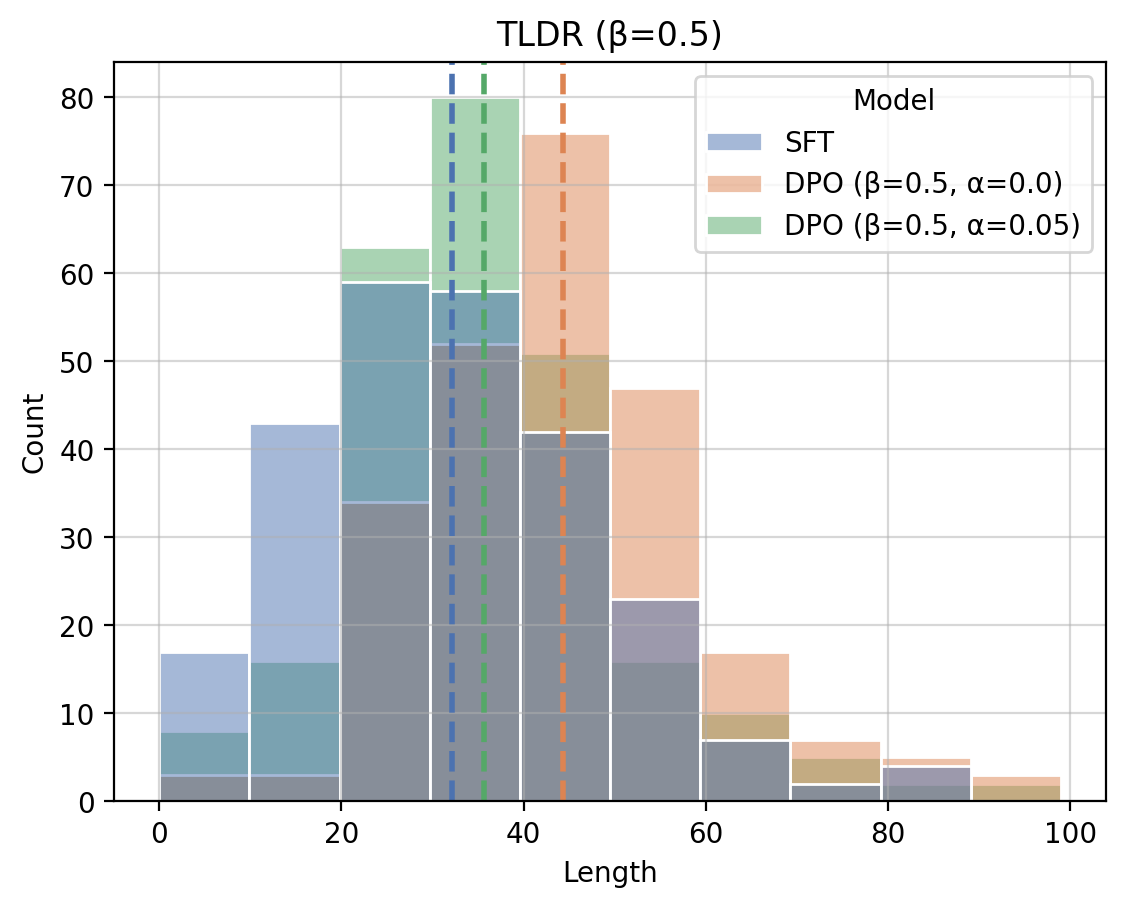}
    
    \includegraphics[width=0.225\textwidth]{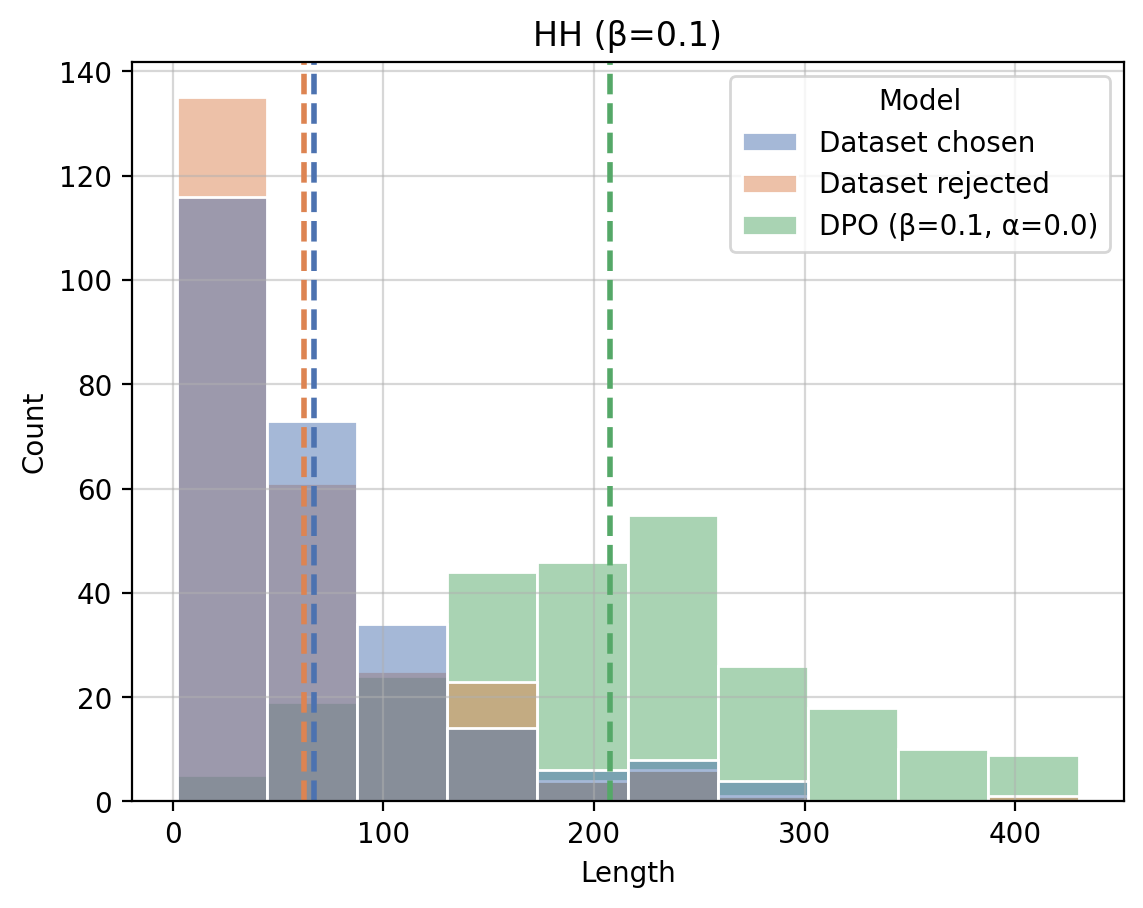}
    \includegraphics[width=0.225\textwidth]{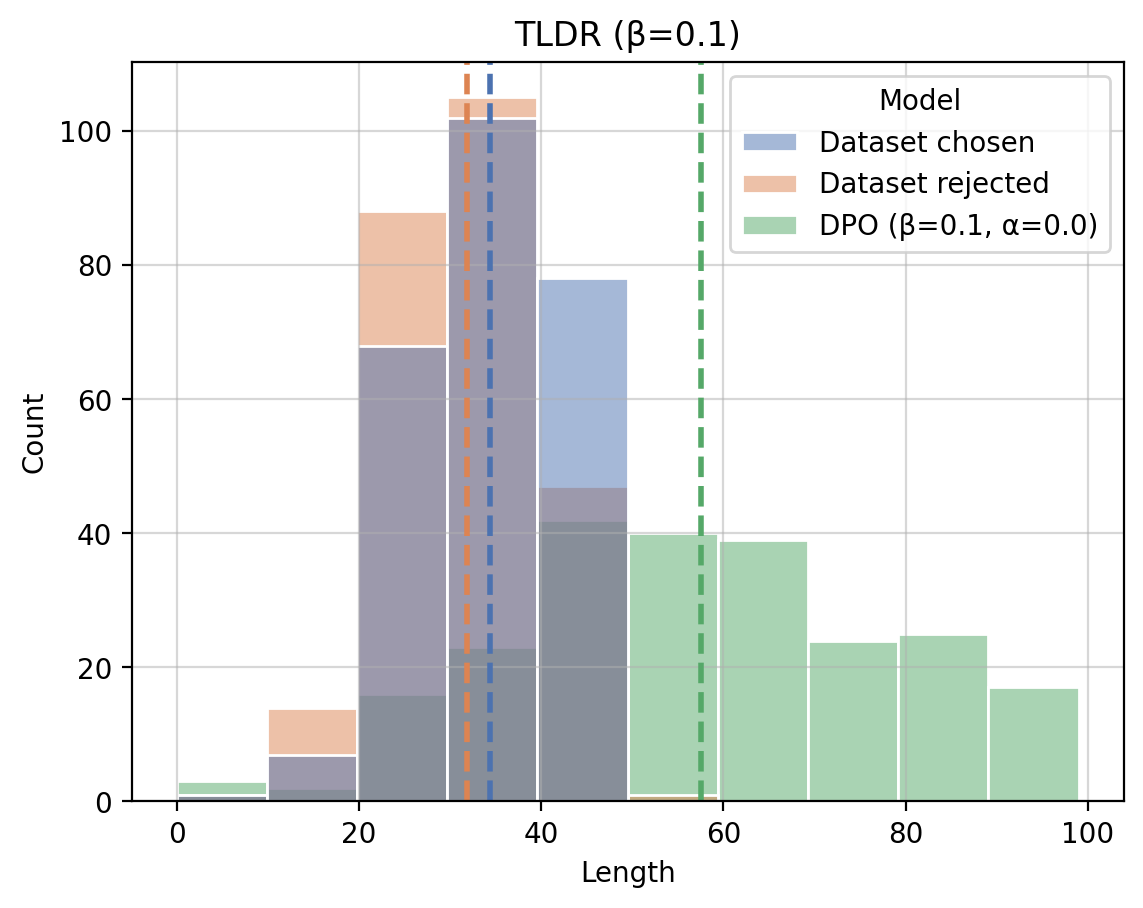}
    \includegraphics[width=0.225\textwidth]{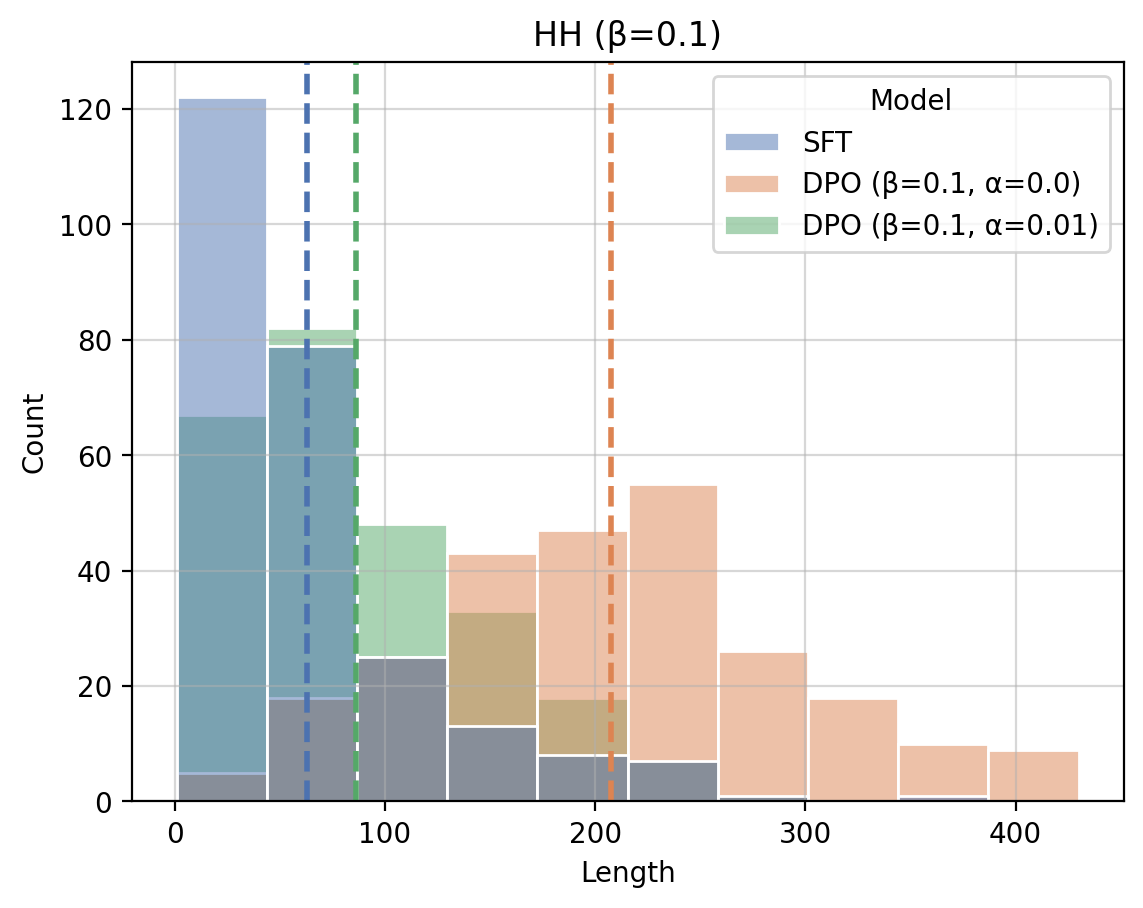}
    \includegraphics[width=0.225\textwidth]{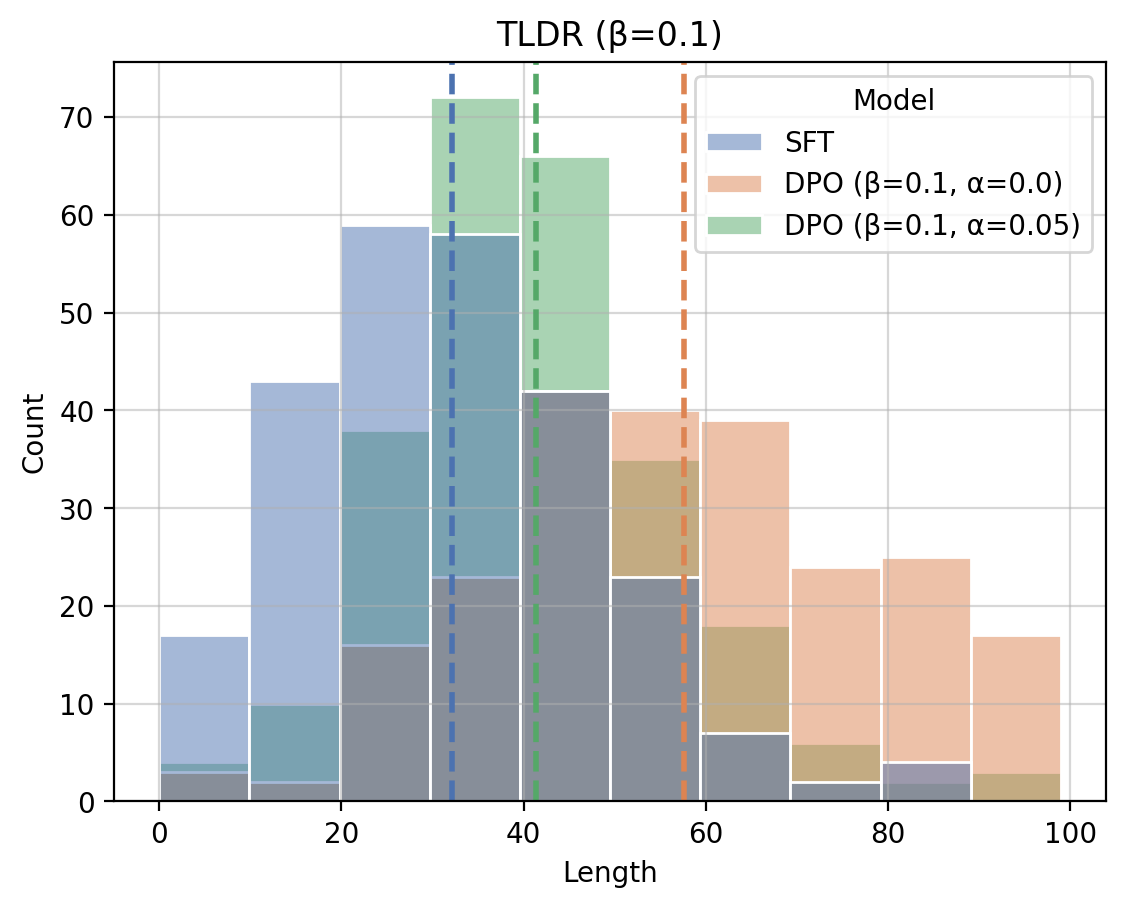}

    \includegraphics[width=0.225\textwidth]{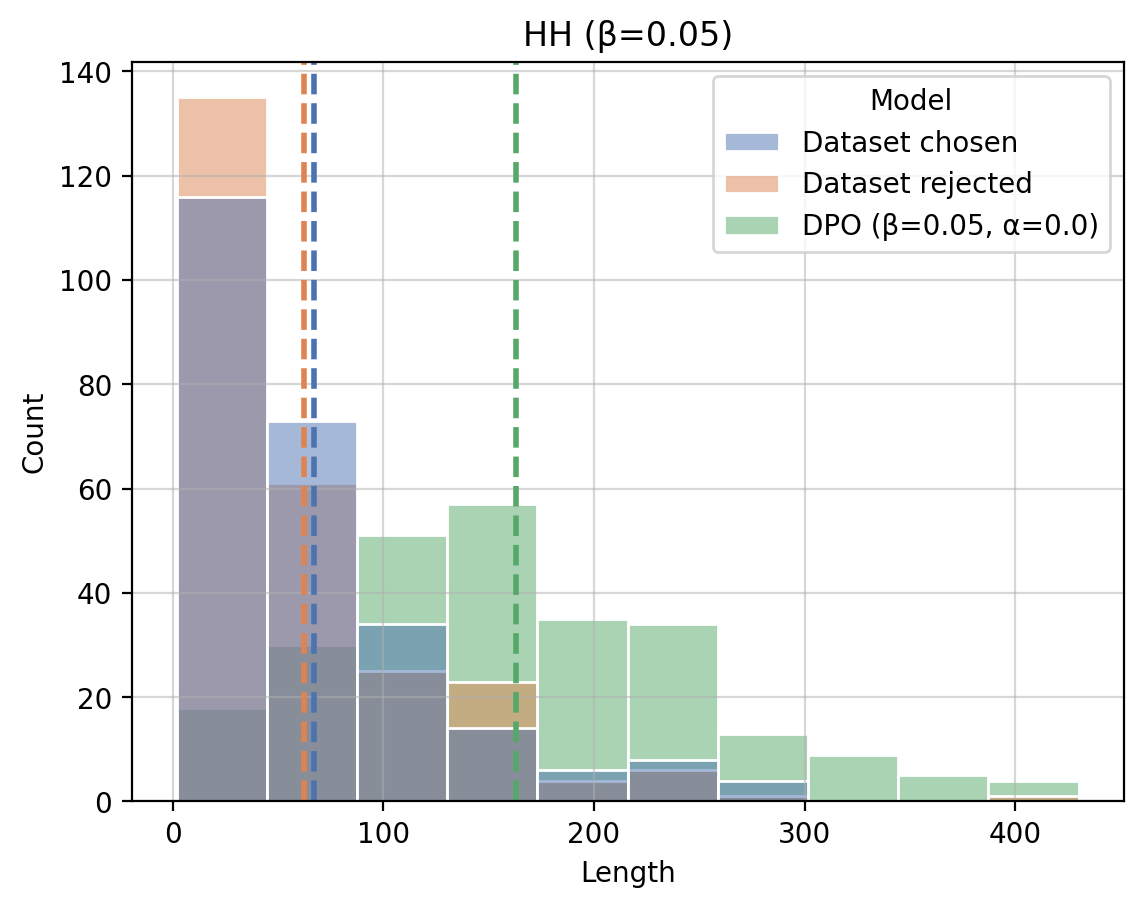}
    \includegraphics[width=0.225\textwidth]{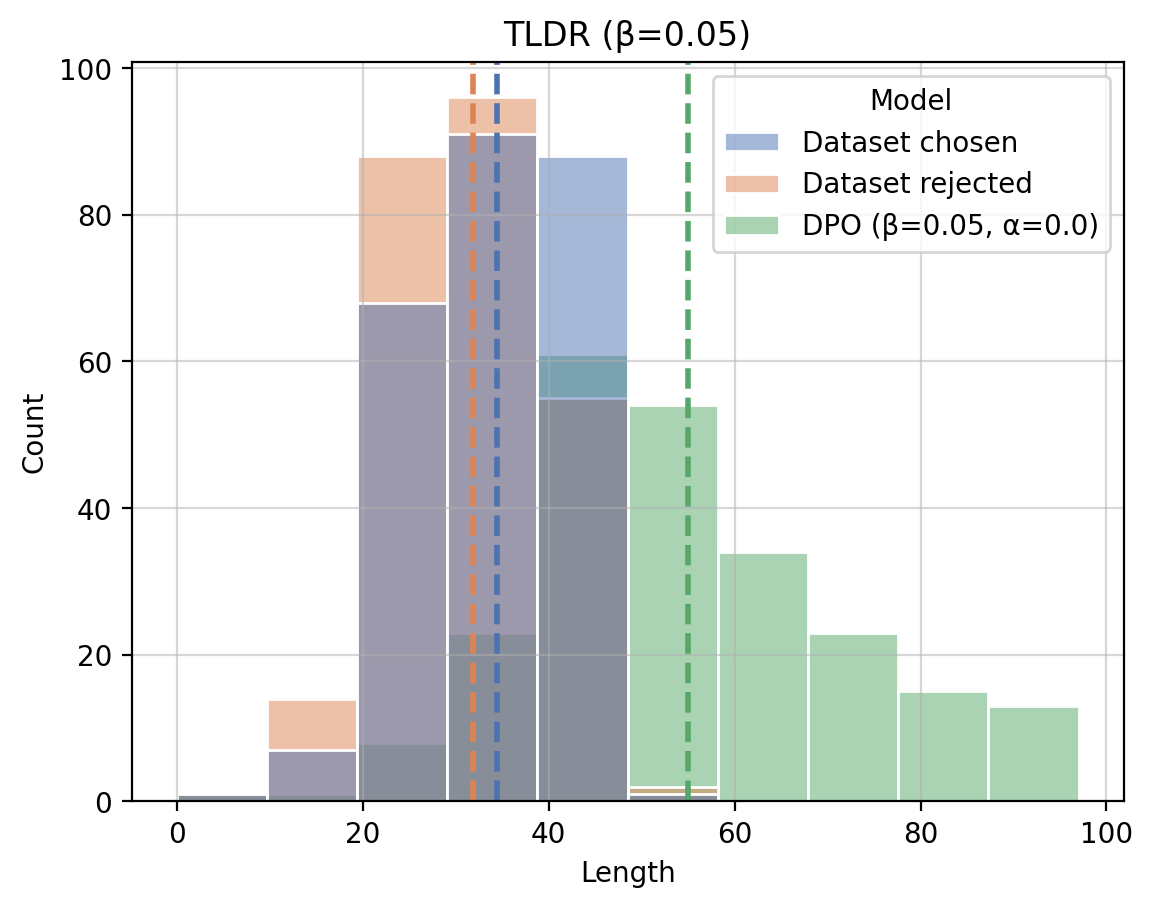}
    \includegraphics[width=0.225\textwidth]{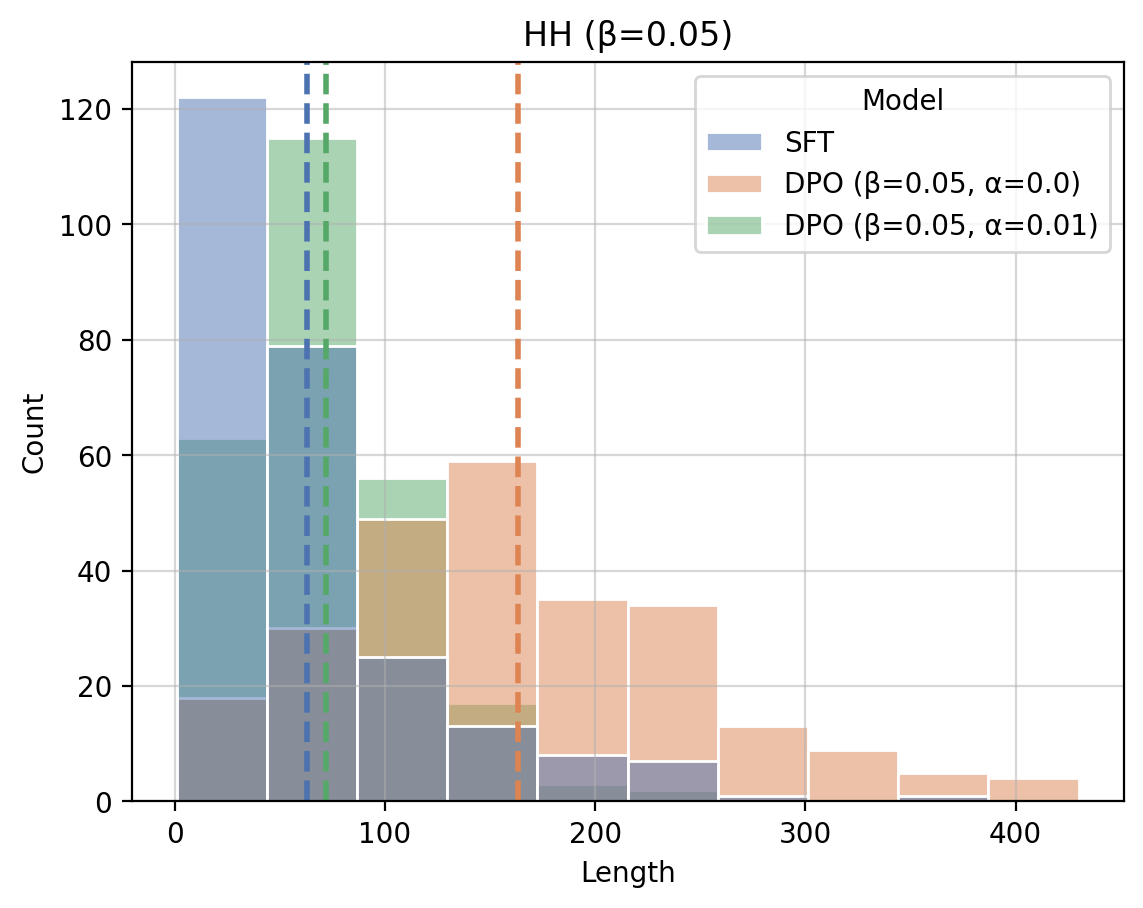}
    \includegraphics[width=0.225\textwidth]{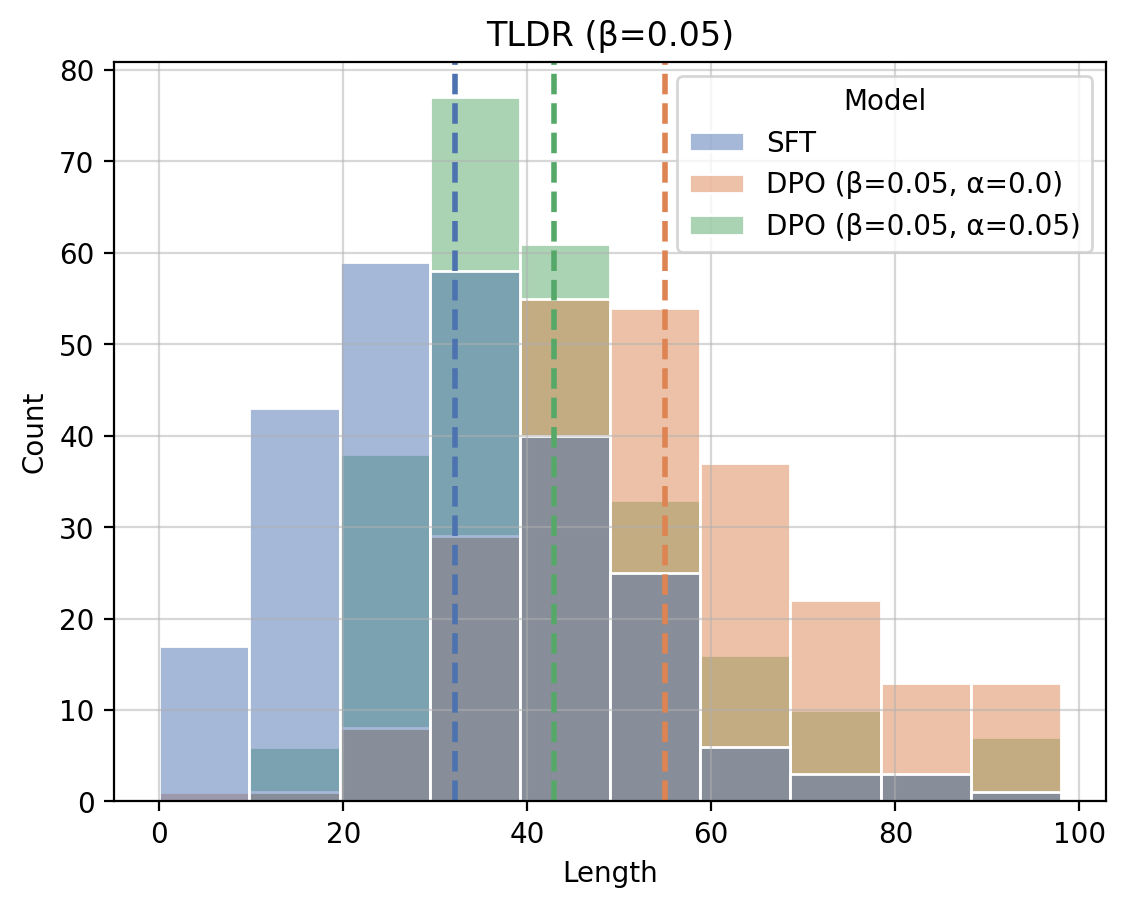}

    \caption{Distribution of response lengths of human feedback datasets, average length is marked by the dashed line. \textbf{First Column:} Statistics on Anthropic's Helpful and Harmless dialogue dataset \cite{bai2022constitutional}. \textbf{Second Column:} Statistics on the Reddit TL;DR summarization dataset \cite{stiennon2022learning}. While both datasets exhibit a small bias in preference towards longer responses, the un-regularized DPO model produces answers twice as long on average, with lengths significantly out of distribution of the feedback dataset. \textbf{Third and Fourth Columns:} Comparison between the SFT, DPO and length-regularized DPO models on HH and TL;DR respectively. While length-regularized DPO algorithm still generates longer answers on average, it stays closer to the SFT model.}
    \label{fig:length_histograms}
\end{figure*}

Recently, Large Language Models (LLMs) have seen significant improvements in capabilities, such as code-generation, mathematical reasoning, and tool use. Importantly, they can now fluently interact with users and follow their instructions, leading to their widespread adoption. Fine-tuning with Reinforcement Learning from Human Feedback (RLHF) \citep{christiano2017deep, stiennon2022learning} has been a significant component in those advances and is now a standard part of advanced LLM training pipelines \citep{ouyang2022training, bai2022training, touvron2023llama, jiang2024mixtral, geminiteam2023gemini}. Currently, all the leading LLMs deploy some sort of RLHF pipeline \cite{dubois2024alpacafarm, zheng2023judging, liang2023holistic}. The classical approach consists of three-stages. The first stage begins with a general model pre-trained with next-token prediction on a large corpus of text \citep{radford2019language, brown2020language}, which is then further-tuned for instruction-following purposes \cite{wei2022finetuned}. In the second stage, the model is prompted with general requests, and generates multiple possible answers, which are then ranked by the user. These ratings are used to train a reward model, which represents human preferences \citep{christiano2017deep, stiennon2022learning, ziegler2020finetuning, bai2022training, touvron2023llama}. In the final stage, the instruction-tuned LLM is further trained to maximize expected rewards from the previously trained reward model  (a proxy for user preferences) using general purpose reinforcement learning algorithms \citep{schulman2017proximal, mnih2016asynchronous}. While successful, this pipeline is technically complex and computationally expensive, mainly due to the final stage of RL optimization.

The quality of the learned reward model is crucial for the RLHF process \cite{touvron2023llama}. However, prior works have demonstrated that reward models can be exploited \citep{casper2023open, gao2022scaling} due to a Goodhart's law effect \cite{ clark2016faulty, manheim2019categorizing, skalse2022defining, lambert2023alignment}. Under this phenomenon, the model can achieve high rewards during the RL training while generating undesirable behaviours \cite{gao2022scaling, dubois2024alpacafarm}. A particular case of the reward exploitation phenomenon is the well-known verbosity issue - models fine-tuned with RLHF generate significantly longer answers, without necessarily improving the actual quality \citep{singhal2023long, kabir2023answers}. This has been linked to an explicit bias in the preference data towards longer responses \cite{singhal2023long}. However, the statistical increase in verbosity of RLHF-trained models significantly outmatches the the difference of distribution lengths between the preferred and rejected answers. This effect is even observed in in strong propriety models, such as GPT-4 \cite{openAI2022}, which is now frequently used to evaluate the performance of other LLMs \cite{dubois2024alpacafarm, zheng2023judging, zeng2023llmbar}. However, even as an evaluator, GPT-4 exhibits strong preferences for length. Prior work \cite{wang2023far} has noted that when evaluating 13B parameter models in head-to-head comparisons with the Davinci-003 model, win rates and the average number of unique tokens in the model's response have correlation of 0.96. 

Recently, Direct Preference Optimization \cite{rafailov2023direct} has emerged as an alternative to the standard RLHF pipeline. The key observation of DPO is that the reward model can directly be re-parameterized through the optimal LLM policy obtained in the reinforcement learning stage. This allows us to directly train the language model through the reward learning pipeline, eliminating the need for the reinforcement learning stage. This algorithm has become widely used, since it can train completely offline, yielding better simplicity of tuning, speed, and resource efficiency, while still maintaining performance \citep{dubois2024alpacafarm, jiang2024mixtral}. For these reasons, it has also been widely adopted by the open-source community. At the time of this writing, 9 out of the top 10 models on the HuggingFace Open LLM Leaderboard use DPO as part of their training pipeline. 

While the question of length exploitation has been extensively studied in the classical RLHF pipeline, it has not been explored in the DPO setting. RLHF-style reward models are explicit, making them susceptible to issues like reward overoptimization \cite{gao2022scaling}. It is unclear whether these issues transfer to DPO, where the lack of an explicit reward model means the problem of reward overoptimization is harder to define. To complicate this issue, others have argued that apparent gains in open-source model performance across automated benchmarks are driven by evaluator's verbosity bias  \cite{tweet2024Liu}. These statistics are demonstrated in Figure \ref{fig:length-rank}, as open-source models can match the overall performance of proprietary ones, but lag significantly on length-corrected basis.

\noindent\textbf{We make several contributions in our work:} First, we show the length exploitation is quite prevalent in DPO. We demonstrate empirically (for the first time) that in this settings OOD extrapolation issues emerge similarly to classical RLHF. Next, we derive a simple but efficient regularization approach, showing it can effectively control verbosity and minimally impact performance even under a length-biased judge which also explains other empirical phenomena in DPO training, such as early convergence. 

\section{Preliminaries}
In this section, we will outline the core components of the standard RLHF pipeline \citeauthor{ziegler2020finetuning, stiennon2022learning, bai2022training, ouyang2022training}) and the Direct Preference Optimization algorithm \cite{rafailov2023direct}, which is central to our analysis and regularization derivations.

\subsection{Reinforcement Learning From Human Feedback}\label{sec:rlhf_pipeline}
The standard RLHF pipeline consists of three stages: 1) we first pre-train a general LLM for instruction-following purposes with supervised fine-tuning (SFT); 2) next, we gather human feedback and train a parameterized reward model; 3) we further optimize the LLM in a reinforcement learning loop using the trained reward model.

\noindent\textbf{SFT}: During this stage, we use a dataset of prompts $\mathbf{x}$ and high-quality answers $\mathbf{y}$ to train an LLM with next-token prediction to obtain a model $\pi_\text{SFT}{(\mathbf{y}|\mathbf{x})}$. In our notation, we treat the entire prompt and answer strings as a single variable. 

\noindent\textbf{Reward modeling Phase}: In the second phase the instruction-tuned model is given prompts $\mathbf{x}$, and produces pairs of answers $(\mathbf{y}_1, \mathbf{y}_2)\sim \pi_\text{SFT}{(\mathbf{y}|\mathbf{x})}$. Users then rank the answers. We denote these preferences as $\mathbf{y}_w\succ \mathbf{y}_l \mid \mathbf{x}$, where $\mathbf{y}_w$ and $\mathbf{y}_l$ are the preferred and dispreferred answer. The rankings are usually assumed to be generated by the Bradley-Terry (BT) \cite{bradley1952rankanalysis}, in which the preference distribution $p$ is assumed to be driven by an unobserved latent reward $r(\mathbf{x}, \mathbf{y})$ and the following parametrization:
\begin{equation}\label{eq:bradley-terry}
    p(\mathbf{y}_1\succ \mathbf{y}_2 \mid x)=\frac{\exp\left(r(\mathbf{x}, \mathbf{y}_1)\right)}{\exp\left(r(\mathbf{x}, \mathbf{y}_1)\right) + \exp\left(r(\mathbf{x}, \mathbf{y}_2)\right)}.
\end{equation}
Then, given a dataset of user rankings $\mathcal{D}=\bigl\{\mathbf{x}^{(i)}, \mathbf{y}_w^{(i)}, \mathbf{y}_l^{(i)}\bigr\}_{i=1}^N$, we can train a parameterized reward model $r_{\phi}(\mathbf{x},\mathbf{y})$ using maximum likelihood: 
\begin{align}\label{eq:reward_model}
   & \mathcal{L}_R(r_{\phi}, \mathcal{D}) = \nonumber\\-& \mathbb{E}_{(\mathbf{x}, \mathbf{y}_w, \mathbf{y}_l)\sim \mathcal{D}}\bigl[\log \sigma(r_{\phi}(\mathbf{x}, \mathbf{y}_w)- r_{\phi}(\mathbf{x}, \mathbf{y}_l))\bigr]
\end{align}
where $\sigma$ is the logistic function.

\noindent\textbf{Reinforcement Learning Phase}: During the final phase, we use the learned reward function in an RL loop where the LLM is treated as a policy. The most common optimization objective is the following:
\begin{align}\label{eq:RL}
\max_{\pi_{\theta}}  \mathbb{E}_{\mathbf{x}\sim \mathcal{D}, \mathbf{y}\sim \pi_{\theta}(\mathbf{y} \mid \mathbf{x})}\bigl[ r_{\phi}(\mathbf{x}, \mathbf{y})\bigr] - \nonumber\\ \beta\mathbb{D}_{\textrm{KL}}\bigl[\pi_{\theta}(\mathbf{y}\mid \mathbf{x})\mid \mid \pi_\text{ref}{(\mathbf{y}|\mathbf{x})}\bigr]
\end{align}
where $\pi_\text{ref}{(\mathbf{y}|\mathbf{x})}$ is a reference distribution (usually taken to be $\pi_\text{ref}{(\mathbf{y}|\mathbf{x})}$) and $\beta$ is a hyper-parameter. This objective trades off maximizing the reward $r_{\phi}(\mathbf{x},\mathbf{y})$ and the regularizing divergence term, which prevents the policy from drifting far from $\pi_\text{ref}{(\mathbf{y}|\mathbf{x})}$. This objective is then optimized using a general purpose RL algorithm, such as PPO \cite{schulman2017proximal}.

\subsection{Direct Preference Optimization}\label{sec:DPO}

Direct Preference Optimization \cite{rafailov2023direct} starts with the same objective as Eq. \ref{eq:RL}. However, DPO assumes we have access to the ground truth reward $r(\mathbf{x},\mathbf{y})$ and derives an analytical transformation between the optimal reward and optimal policy. This can be substituted back into the reward optimization objective in Eq. \ref{eq:reward_model}, which allows us to train the optimal model directly on the feedback data using the following objective:
\begin{align}\label{eq:DPO}
\mathcal{L}_{\mathrm{DPO}}\left(\pi_\theta ; \pi_{\mathrm{ref}}\right)= \nonumber\\-\mathbb{E}_{\left(\mathbf{x}, \mathbf{y}_w, \mathbf{y}_l\right) \sim \mathcal{D}}\Big[\log \sigma\Big(&\beta \log \frac{\pi_\theta\left(\mathbf{y}_w \mid \mathbf{x}\right)}{\pi_{\mathrm{ref}}\left(\mathbf{y}_w \mid \mathbf{x}\right)} -\nonumber\\
&\beta \log \frac{\pi_\theta\left(\mathbf{y}_l \mid \mathbf{x}\right)}{\pi_{\mathrm{ref}}\left(\mathbf{y}_l \mid \mathbf{x}\right)}\Big) \Big]
\end{align}
Here, the parameter $\beta$ is the same as in Eq. \ref{eq:RL}, and similarly controls the trade-off between expected reward and divergence from the model initialization. The DPO objective is attractive since it allows us to recover the optimal model using a standard classification loss, without the need for on-policy sampling or significant amount of hyper-parameter tuning.
Eq. \ref{eq:DPO} resembles the reward modeling objective in Eq. \ref{eq:reward_model} under the parameterization

\begin{equation}\label{eq:implicit_reward}
r_{\theta}(\mathbf{x}, \mathbf{y}) = \beta \log \frac{\pi_\theta\left(\mathbf{y} \mid \mathbf{x}\right)}{\pi_{\mathrm{ref}}\left(\mathbf{y} \mid \mathbf{x}\right)}
\end{equation}

We will refer to this as the DPO "implicit reward". Theorem 1 in \cite{rafailov2023direct} shows that this is indeed a valid parameterization of a reward model without loss of generality. If we substitute this form of $r_{\theta}(\mathbf{x}, \mathbf{y})$ into the RL objective \ref{eq:RL}, we can obtain the optimal solution in a closed form, which happens to be $\pi_{\theta}$. We will return to the interpretation of DPO as an implicit reward function later on in our analysis of out-of-distribution bootstrapping. 

\section{Building in Explicit Regularization in DPO}\label{sec:R-DPO}

Prior works have explicitly considered length-regularization in the classical RLHF pipeline \cite{singhal2023long}, however these methods do not transfer directly to direct alignment algorithms, such as DPO \cite{rosset2024direct}. We derive a length-regularized version of the algorithm by adding a regularized term to Eq. \ref{eq:RL}. The below considerations hold for a general regularizer, but we focus on a length term $\alpha |y|$, where $\alpha$ is a hyper-parameter and $|y|$ denotes the token-length of the answer $y$. We then formulate the regularized RL problems in the following objective:
\begin{align}\label{eq:regRL}
\max_{\pi_{\theta}}  \mathbb{E}_{\mathbf{x}\sim \mathcal{D}, \mathbf{y}\sim \pi_{\theta}(\mathbf{y} \mid \mathbf{x})}\bigl[r(\mathbf{x}, \mathbf{y})\bigr] - \alpha |\mathbf{y}|-\nonumber\\ \beta\mathbb{D}_{\textrm{KL}}\bigl[\pi_{\theta}(\mathbf{y}\mid \mathbf{x})\mid \mid \pi_\text{ref}{(\mathbf{y}|\mathbf{x})}\bigr]
\end{align}
where we assume that $r(\mathbf{x}, \mathbf{y})$ is still the same latent reward driving human preferences. We can follow the same derivations in \cite{rafailov2023direct} for the reward function $r(\mathbf{x}, \mathbf{y})\bigr] - \alpha |\mathbf{y}|$ and obtain the optimal solution to Eq. \ref{eq:regRL} as
\begin{equation}
    \pi^*(\mathbf{y}|\mathbf{x}) = \frac{1}{Z(\mathbf{x})}\pi_{\mathrm{ref}} e^{\frac{1}{\beta}(r(\mathbf{x},\mathbf{y})-\alpha|\mathbf{y}|)}
\end{equation}
where $Z(\mathbf{x}) = \sum_{\mathbf{y}}\pi_{\mathrm{ref}}e^{\frac{1}{\beta}(r(\mathbf{x},\mathbf{y})-\alpha|\mathbf{y}|)}$. With some simple algebra, we can then obtain the equivalent regularized reward re-formulation:
\begin{equation}\label{eq:exterinsic}
    r(\mathbf{x}, \mathbf{y})=\beta \log \frac{\pi^{*}(\mathbf{y}|\mathbf{x})}{\pi_\text{ref}{(\mathbf{y}|\mathbf{x})}} + \beta \log Z(\mathbf{x}) + \alpha|\mathbf{y}|
\end{equation}
We can then plug in Eq. \ref{eq:exterinsic} into the reward modeling stage in Eq. \ref{eq:reward_model}, which yields the following regularized DPO objective:
\begin{align}\label{eq:rDPO}
\mathcal{L}_{\mathrm{R-DPO}}\left(\pi_\theta ; \pi_{\mathrm{ref}}\right)= \nonumber\\-\mathbb{E}_{\left(\mathbf{x}, \mathbf{y}_w, \mathbf{y}_l\right) \sim \mathcal{D}}\Big[\log \sigma\Big(&\beta \log \frac{\pi_\theta\left(\mathbf{y}_w \mid \mathbf{y}\right)}{\pi_{\mathrm{ref}}\left(\mathbf{y}_w \mid \mathbf{x}\right)} -\nonumber\\
&\beta \log \frac{\pi_\theta\left(\mathbf{y}_l \mid \mathbf{x}\right)}{\pi_{\mathrm{ref}}\left(\mathbf{y}_l \mid \mathbf{x}\right)} + \nonumber\\
&(\alpha|\mathbf{y}_w|-\alpha|\mathbf{y}_l|)
\Big) \Big]
\end{align}
This is similar to the standard DPO objective, except for an additional regularization term ($\alpha|\mathbf{y}_w|-\alpha|\mathbf{y}_l|$) in the logit of the binary classification loss. Recent work \cite{zhou2023onepreferencefitsall}, aims to develop a multi-objective optimization approach and arrives at a similar objective involving a mixture of reward models, while we focus on optimization robustness of DPO. Another concurrent work \cite{chen2024odin} also consider the length exploitation problem in the classical RLHF pipeline, suggesting a similar regularization in the reward modeling stage in Eq. \ref{eq:reward_model}. This regularizer helps disentangle answer quality from length, demonstrating meaningful improvement in length-controlled performance. Our derivations can be seen as the DPO implicit reward counterpart to this classical RLHF approach, explicitly linking the regularized reward modeling problem to an equivalent regularized RL setup.

\begin{table*}[t]
\centering
\begin{tabular}{lccccccc}
    \toprule
    \textbf{Dataset} &
      \multicolumn{3}{c}{\textbf{Preferred Length}} &
      \multicolumn{3}{c}{\textbf{Dispreferred Length}}  \\
      & {Mean} & {Median} & {Std.} & {Mean} & {Median} & {Std.} \\
    \cmidrule(r){2-4}\cmidrule(l){5-7}
    Anthropic RLHF HH & \textbf{79.6} & \textbf{57.0} & \textbf{74.0} & 75.7 & 51.0 & 73.3 \\
Reddit TL;DR & \textbf{37.9} & \textbf{36.0} & \textbf{13.9} & 35.2 & 34.0 & 13.4 \\

    \bottomrule
\end{tabular}
\caption{Summary statistics across preference datasets. Bold indicates maximum between preferred and dispreferred statistic for a particular dataset. Statistics do not exclude long tails.}
\label{table:lengths}
\end{table*}

Similar to the original DPO formulation, the regularized objective still aims to increase the likelihood along the preferred answer, while decreasing the likelihood along the dis-preferred answer, modulated by a weighting term. This term is equivalent to the original DPO formulation with the addition of the regularization margin $\alpha |\mathbf{y}_w| - \alpha |\mathbf{y}_l|$. We can interpret this as an additional per-example learning rate, which up-weighs the gradient on feedback pairs, in which the selected answer is shorter and down-weights the gradient on pairs in which the selected answer is longer, proportional to the difference in length.

\section{Experiments}

In this section we will empirically investigate the verbosity exploitation issues in DPO, the effectiveness of our regularization strategy and the potential causes of these effects. We beging with a description of our evaluation tasks and models. 

\subsection{Datasets and Models}

We utilize three different setups in our experimental setting based on summarization, dialogue and general instruction-following.

\noindent\textbf{Summarization} We use the standard Reddit TL;DR (TL;DR) summarization dataset from \cite{stiennon2022learning}, which consists of a Reddit post and several short summaries, judged for quality and informativeness by human evaluators. 

\noindent\textbf{Dialogue:} For our dialogue experiment we use the Anthropic Helpful and Harmless (HH) datasets \cite{bai2022constitutional}, which consists of general conversations with a language model assistants, which are also ranked by human annotators.

Datasets statistics are included in Table \ref{table:lengths} where exhibit a small length bias in the preferred response. Following \cite{rafailov2023direct}, we use the Pythia 2.8B \cite{biderman2023pythia} for both the dialogue and summarization tasks, carrying out full-parameter fine-tuning using the DPO original codebase\footnote{https://github.com/eric-mitchell/direct-preference-optimization} with default hyperparameters, except when noted otherwise. See \ref{sec:exp_details} for more details.

\subsection{Length Exploitation in DPO and Effectiveness of Regularization}

We first consider the Anthropic Helpful and Harmless and Reddit TL;DR datasets. For both tasks, we train models with three parameter values $\beta\in[0.5, 0.1, 0.05]$ and then sample 256 answers using prompts from the evaluation dataset. The length histograms are shown in Fig. \ref{fig:length_histograms}. The first two columns show the answer length distribution for the set of preferred, rejected and DPO-generated answers, with each row corresponding to a different $\beta$ parameter value. We see that the DPO generated answers are, on average, significantly longer than both the preferred and rejected answers. Models trained with smaller values of $\beta$ generate longer responses on average, which is expected since $\beta$ controls the deviation from the initial policy. Not only does the DPO model generate longer answers, it also generates answers that are significantly out-of-distribution in terms of length from the offline preference dataset.

The third and fourth column in Fig. \ref{fig:length_histograms} show results for the SFT, DPO the length-regularized DPO model introduced in Section \ref{sec:R-DPO}. We use  $\alpha = 0.01$ and $\alpha = 0.05$ for the Anthropic Helpful and Harmless and Reddit TL;DR datasets respectively. While the length-regularized models still show mild increase in average length, they match the SFT model much more closely. Moreover, they do not generate answers with significantly out-of-distribution lengths. This indicates that the proposed algorithm can efficiently regularize the verbosity of the trained model.

\subsection{Length Versus Quality Trade-Offs}

\begin{figure*}[t]
    \centering
    \includegraphics[width=0.475\textwidth]{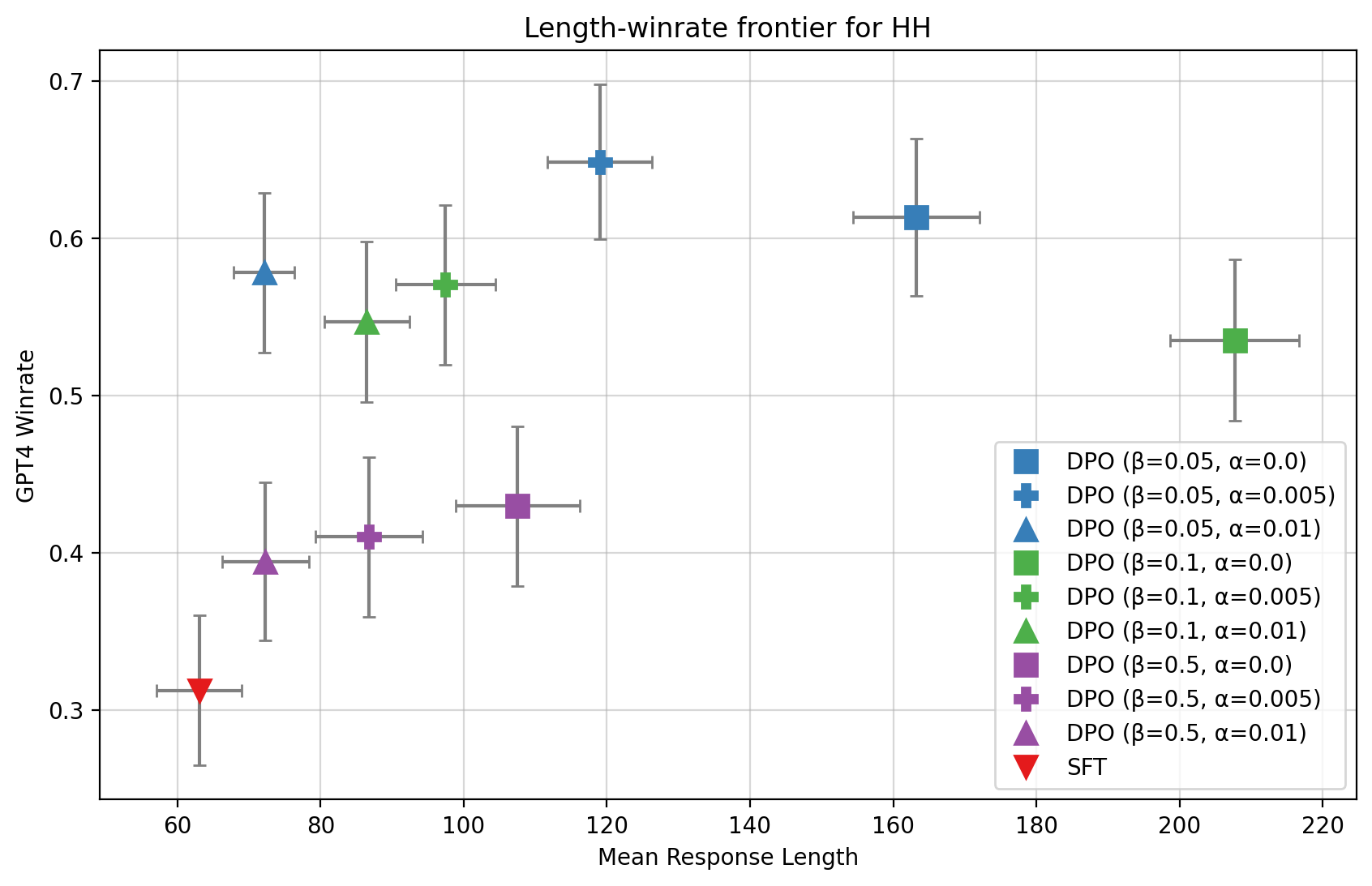}
    \includegraphics[width=0.475\textwidth]{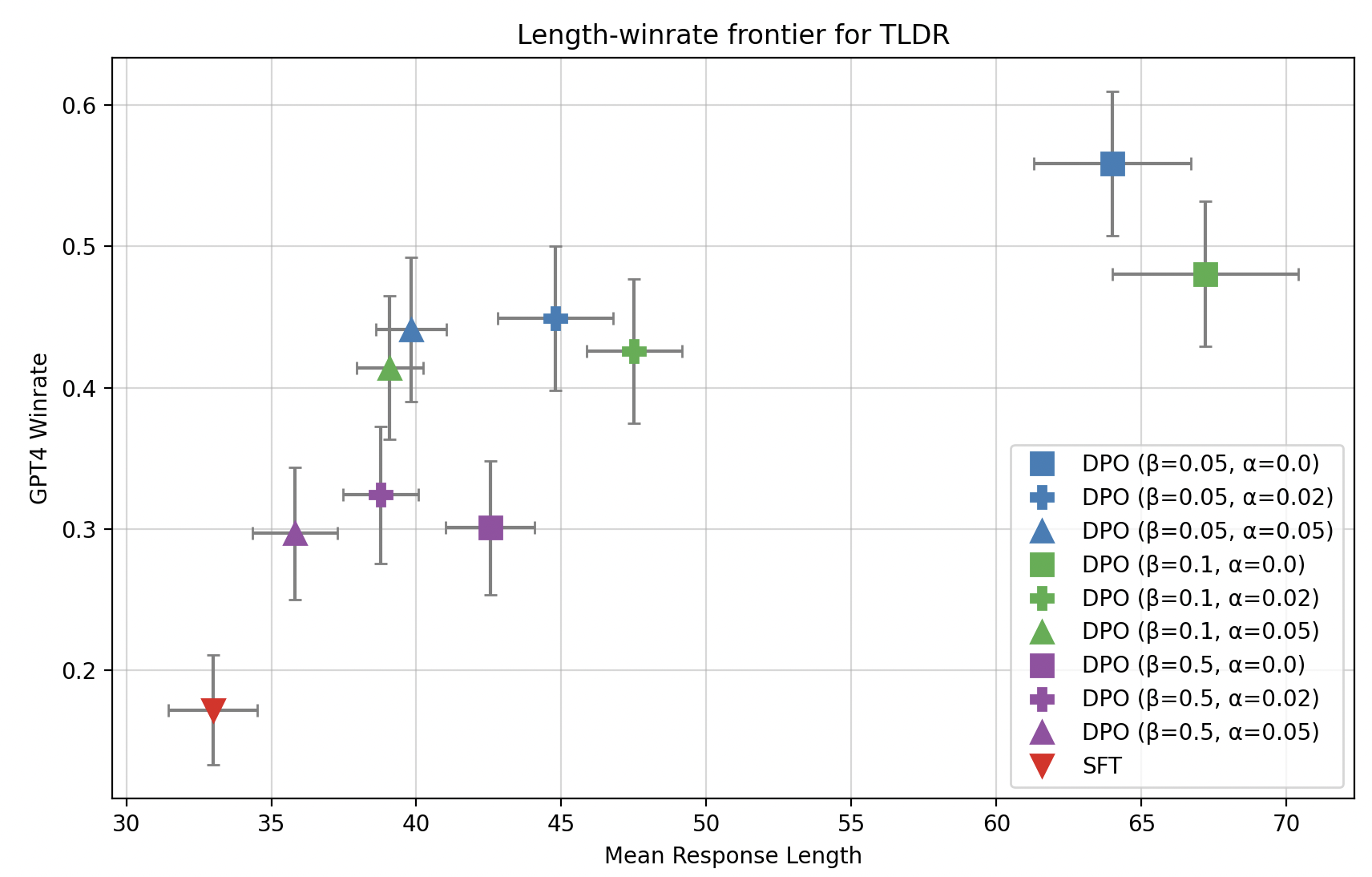}
    \caption{Sampled lengths vs. GPT-4 winrates for HH and TL;DR test sets. 256 samples evaluated for length and winrates. GPT-4-0613 used as judge with prompt similar to \cite{rafailov2023direct}, with random position flipping.}
    \label{fig:len_win}
\end{figure*}

In this section, we evaluate the length versus quality model trade-offs. For the Anthropic Helpful and Harmless and Reddit TL;DR datasets, we use the answers sampled from DPO policies and compare them head-to-head against the dataset preferred answer, with GPT-4 as an evaluator. Our main results are shown in Fig. \ref{fig:len_win}, which plots model win rates against average answer length, with 90\% confidence intervals. We again evaluate all models with $\beta \in [0.05, 0.1. 0.5]$. For HH, we use $\alpha \in [0, 0.005, 0.01]$; for TL;DR, we use $\alpha \in [0, 0.2, 0.5]$ ($\alpha =0$ is standard DPO). Similar to before, we see that the length-regularized training can efficiently control verbosity, significantly decreasing the average length of the answers as compared to standard DPO. Moreover, on the HH task, regularization also leads to mild improvement in win rates, but a slight decrease on TL;DR (although both of these are not statistically significant). These results are quite promising, as GPT-4 is known to have a significant length bias in its preferences \cite{wang2023far, singhal2023long}. On both HH and TL;DR, the length-regularized experiments with $\beta = 0.05$ and $\beta=0.01$ match the average lengths of the corresponding $\beta=0.5$ runs, but achieve statistically significant higher corresponding win rates, with close to 20\% improvement on HH and close to 15\% improvement on TL;DR.

\subsection{Is Length a Proxy for KL-Divergence?}

\begin{figure*}[t]
    \centering
    \includegraphics[width=0.475\textwidth]{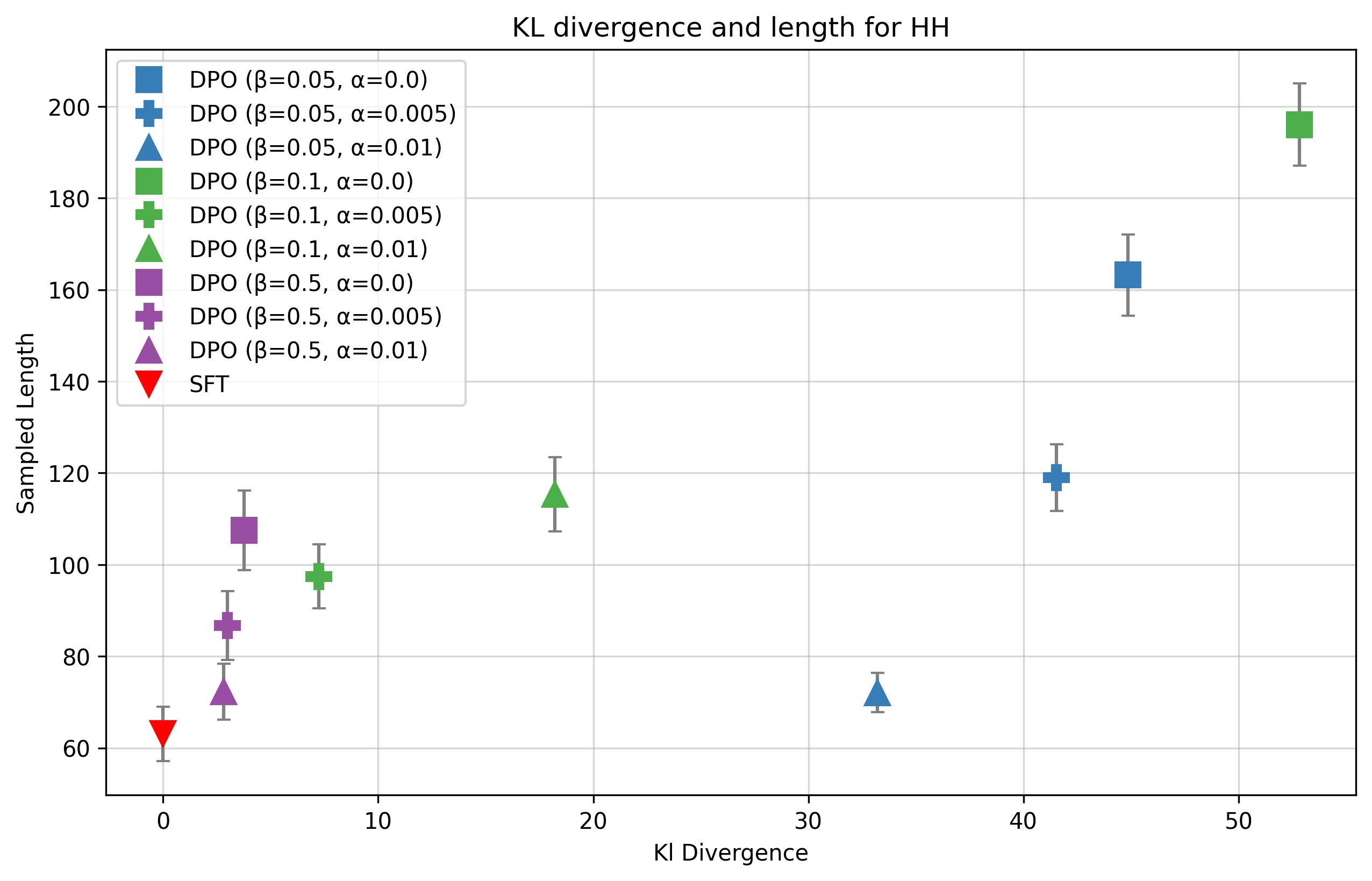}
    \includegraphics[width=0.475\textwidth]{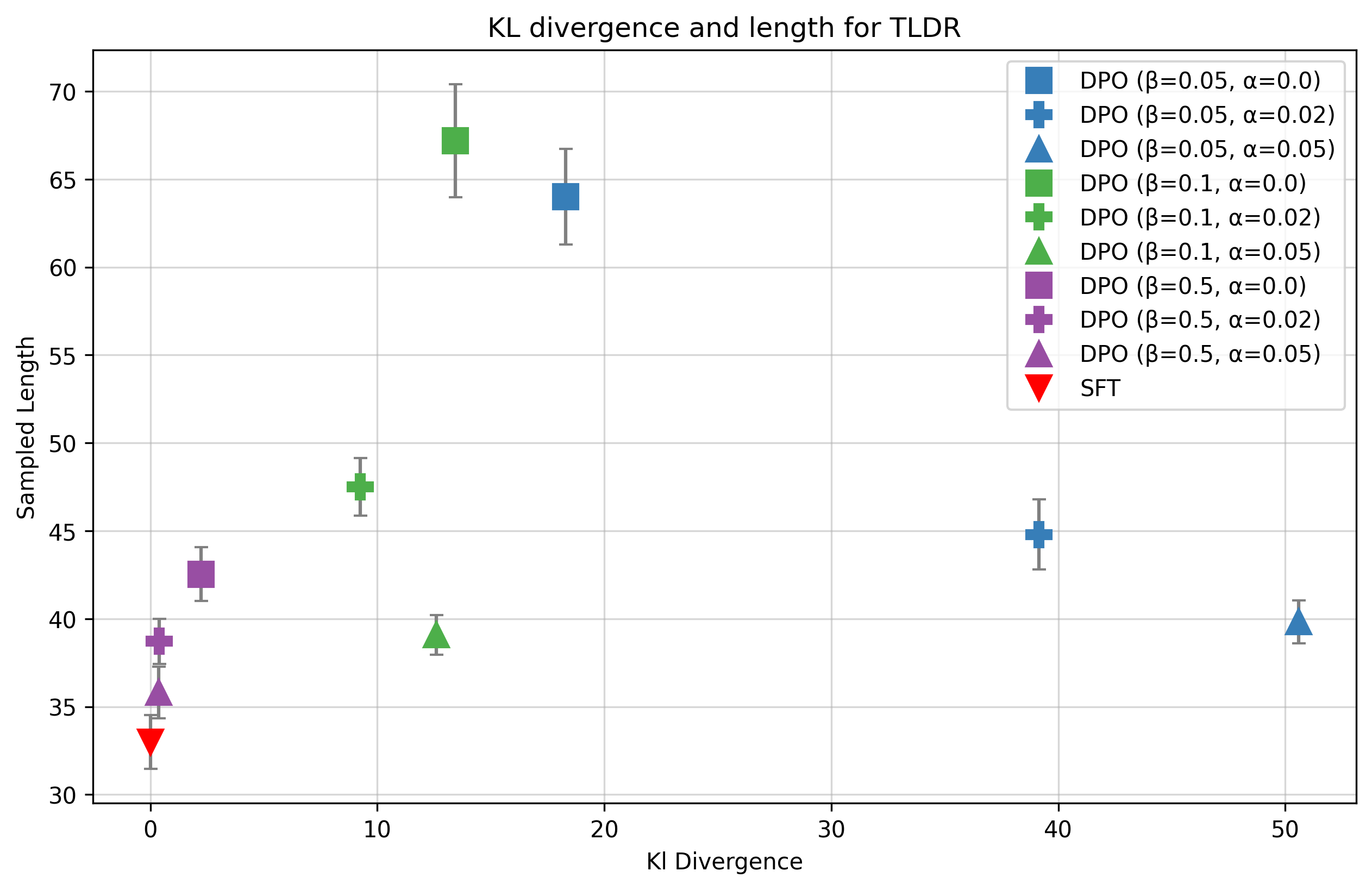}
    \caption{KL divergence vs. sampled lengths for HH and TL;DR. Particularly for TL;DR, the KL budget is only weakly correlated with winrates and length.}
    \label{fig:kl_win}
\end{figure*}

In the constrained RL problem in Eq. \ref{eq:RL} and the corresponding DPO objective in Eq. \ref{eq:DPO}, the $\beta$ parameter controls the degree of policy divergence from the initial reference model. In Fig. \ref{fig:length_histograms} and Fig. \ref{fig:len_win}, we see that average length of the model generated answers is inversely proportional to the $\beta$ parameter. In this section, we investigate the relationship between the length-regularized DPO objective in Eq. \ref{eq:rDPO} and the KL divergence from the initial policy. In Fig. \ref{fig:kl_win}, we plot the trained policy's KL divergence from $\pi_{\mathrm{ref}}$ for the different values of $\beta$ and $\alpha$ parameters. We see only a weak correlation between KL divergence and length. For both HH and TL;DR, length-regularized models trained with $\beta=0.05$ and $\beta=0.01$ match the average length of train runs with $\beta = 0.5$ (Fig. \ref{fig:len_win}). At the same time, these runs have statistically significant higher KL divergences and win rates as shown in Fig. \ref{fig:kl_win}. We hypothesize that this indicates the existence of different factors driving human preference, with length only partially accounting for the policy KL divergence.

\subsection{DPO and Early Convergence}

\begin{figure*}[t]
    \centering
    \includegraphics[width=0.325\textwidth]{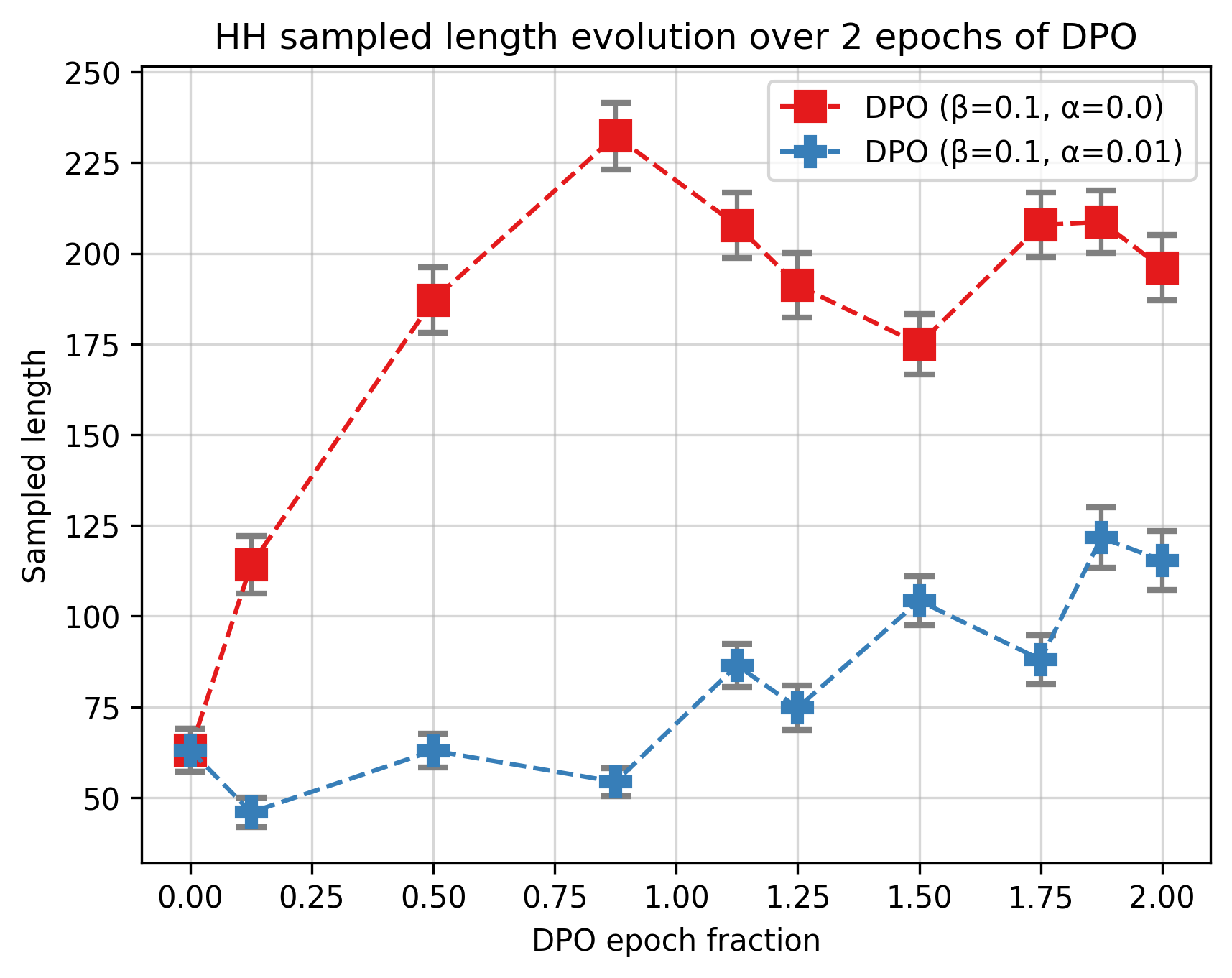}
    \includegraphics[width=0.325\textwidth]{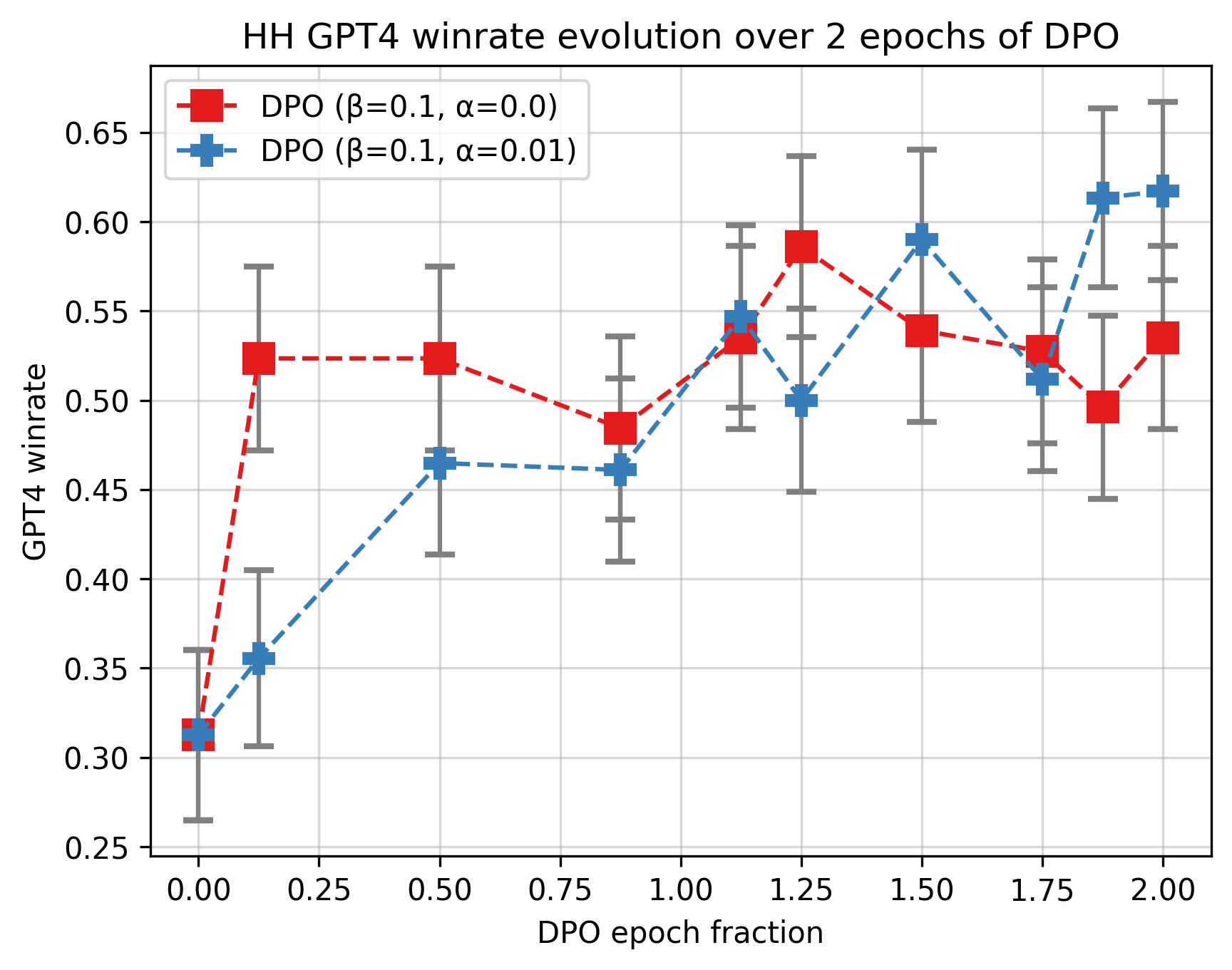}
    \includegraphics[width=0.325\textwidth]{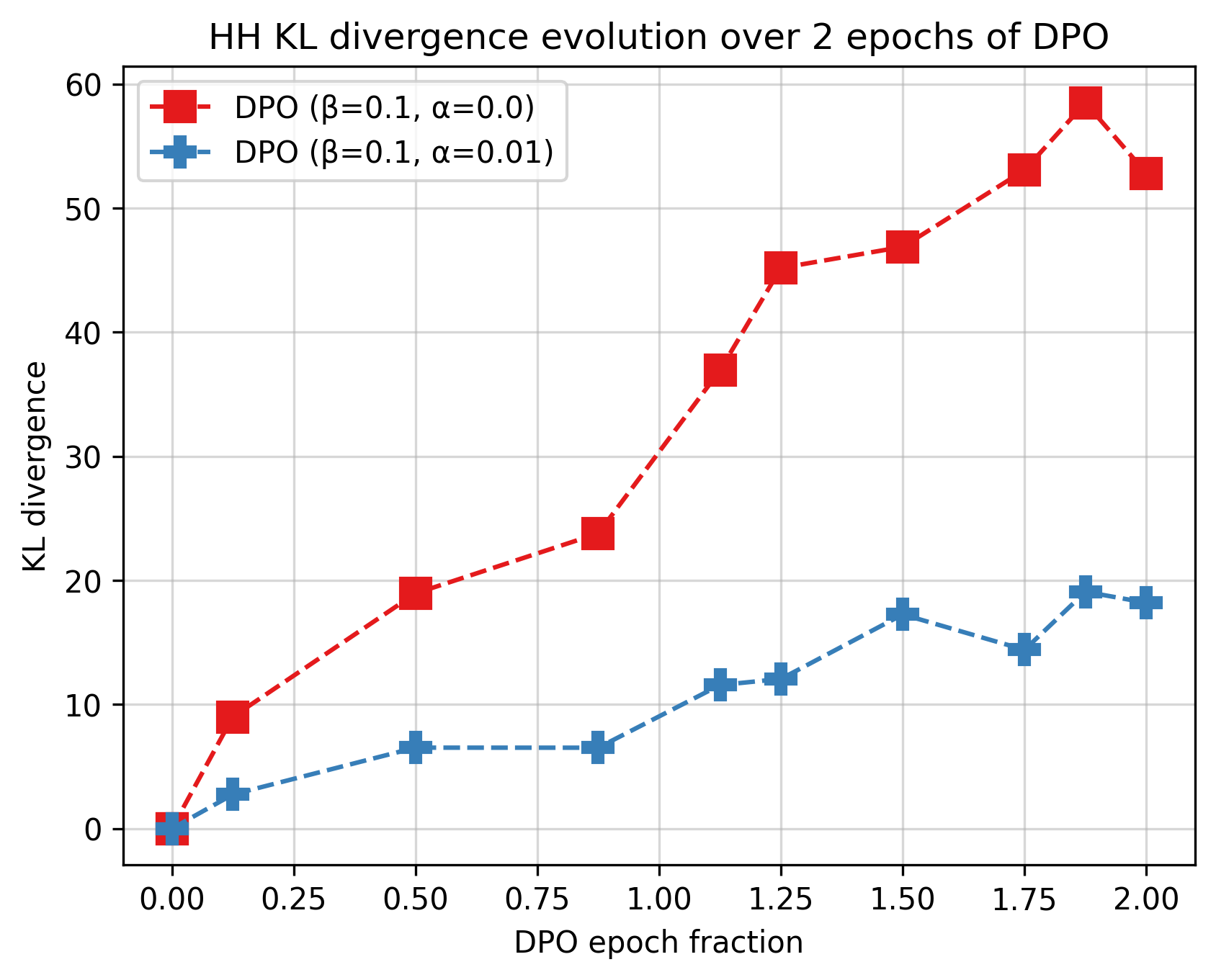}
    \caption{Evolution of HH sample length, winrates, and KL divergence within two epochs of DPO training. Error bars indicate 90\% confidence intervals. The length-regularized model achieves higher final winrates over the regular DPO model, at less than 40\% of the KL budget and almost half the response length. Moreover, the length-regularized model demonstrates steady improvement throughout training, while standard DPO performance peaks early on tin the first epoch and does not improve further, indicating it is not able to learn more complex features of the data.}
    \label{fig:evolution}
\end{figure*}

In \cite{rafailov2023direct}, the authors show early convergence of the DPO algorithm on the HH dataset. DPO achieves its best performance within a few hundred gradient steps, and does not improve with further training. Similar observations have also been made within the open-source community. We claim that this effect is likely due to length exploitation and the biased GPT-4 evaluator. In Fig. \ref{fig:evolution}, we consider the training progression on the HH dataset with $\beta=0.1$. We compare the regular DPO run ($\alpha=0$) with the length-regularized one ($\alpha = 0.1$). We train for two epochs and evaluate intermediate checkpoints on the same set of prompts for average answer length, win rates, and KL divergence. Within the first 10\% of the epoch, the standard DPO run produces answers almost twice as long as the SFT model. Standard DPO achieves its highest win rate here, with only KL divergence and average length increasing steadily with further training. In contrast, the length-regularized run sees little to no intermediate increase in length, but steady improvement in win rates throughout training and slow increases in divergence from the reference policy. Our final regularized checkpoint outperforms the non-regularized model along all fronts (KL, winrate, and length) at the end of 2 epochs. We hypothesize that the regular DPO training quickly increases length, exploiting the evaluator's bias, but fails to capture more complex preference features. On the other hand, the length-regularized training run is able to disentangle the verbosity component and fit other, more difficult quality features over a longer training period.

\subsection{What Drives Length Exploitation?}

\begin{figure*}[t]
    \centering
    \includegraphics[width=0.325\textwidth]{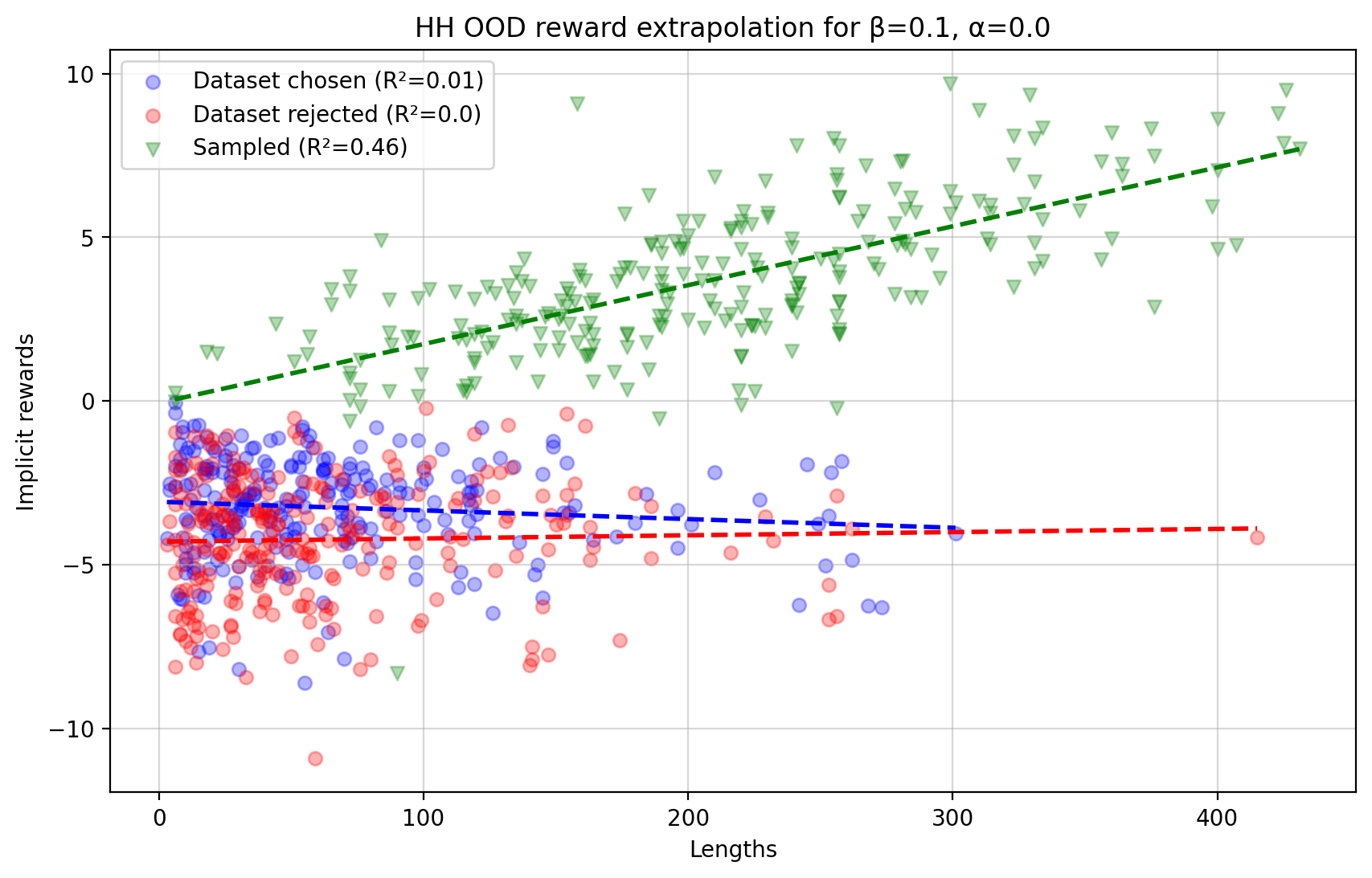}
    \includegraphics[width=0.325\textwidth]{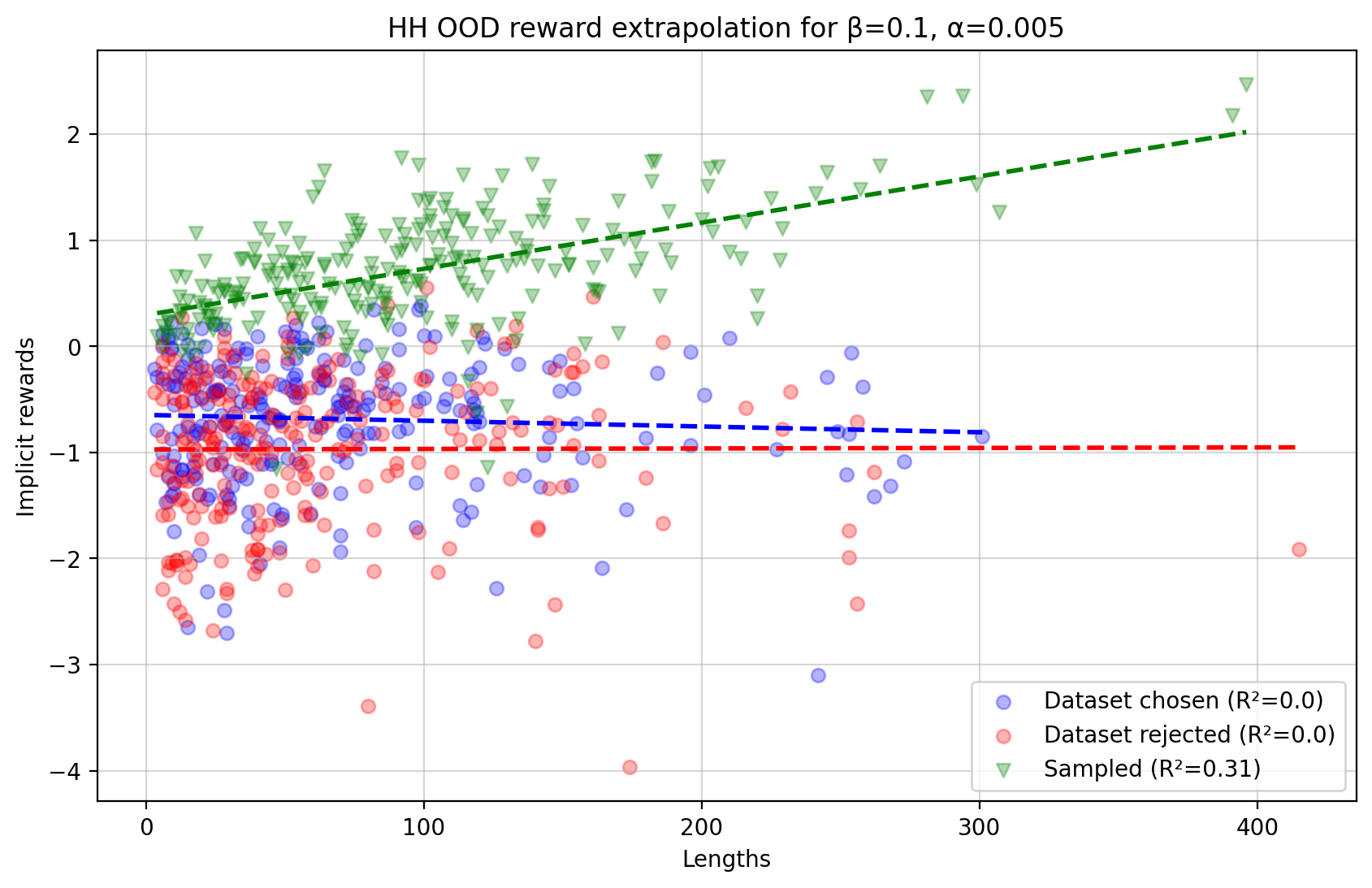}
    \includegraphics[width=0.325\textwidth]{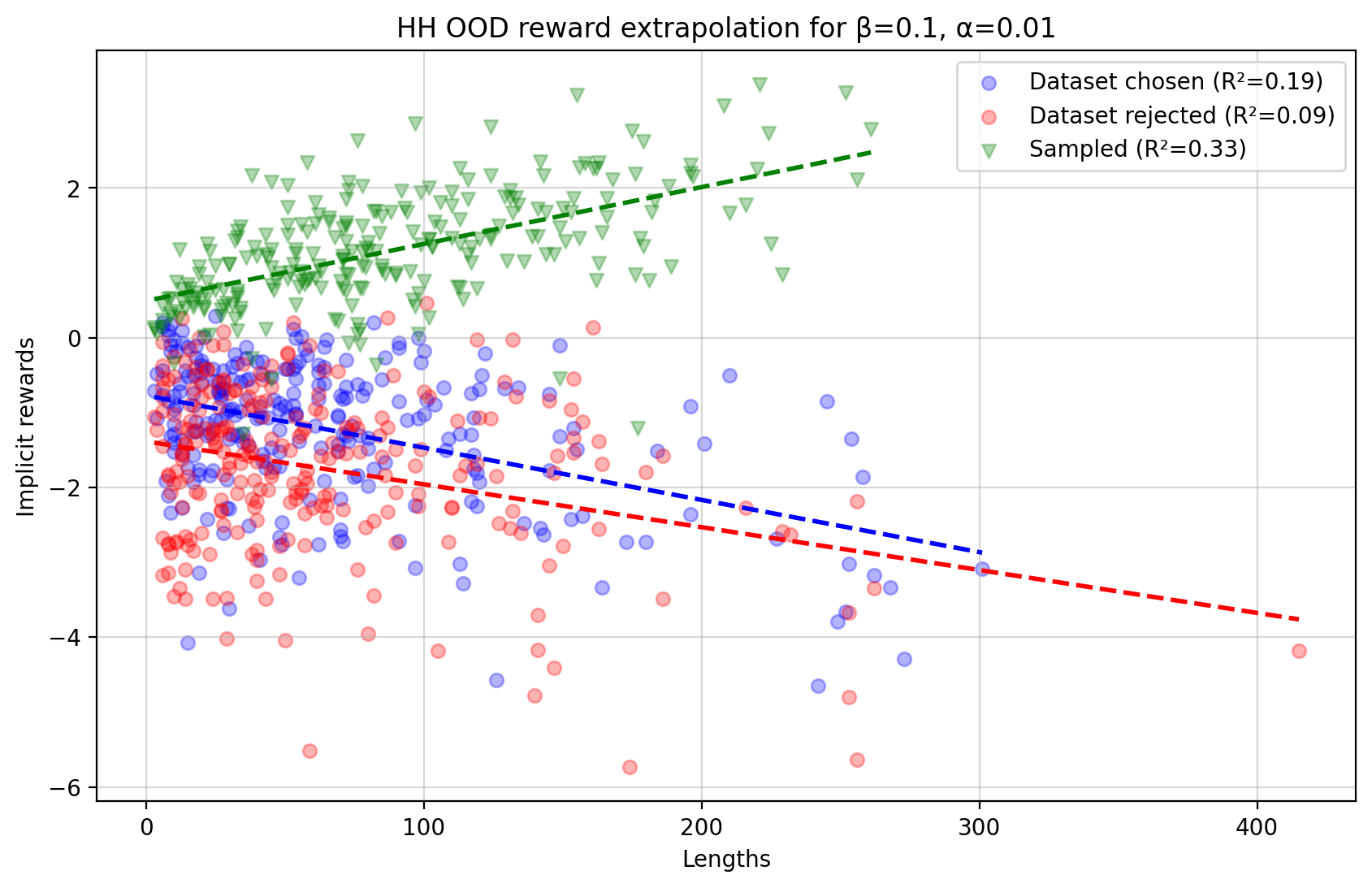}
    \includegraphics[width=0.325\textwidth]{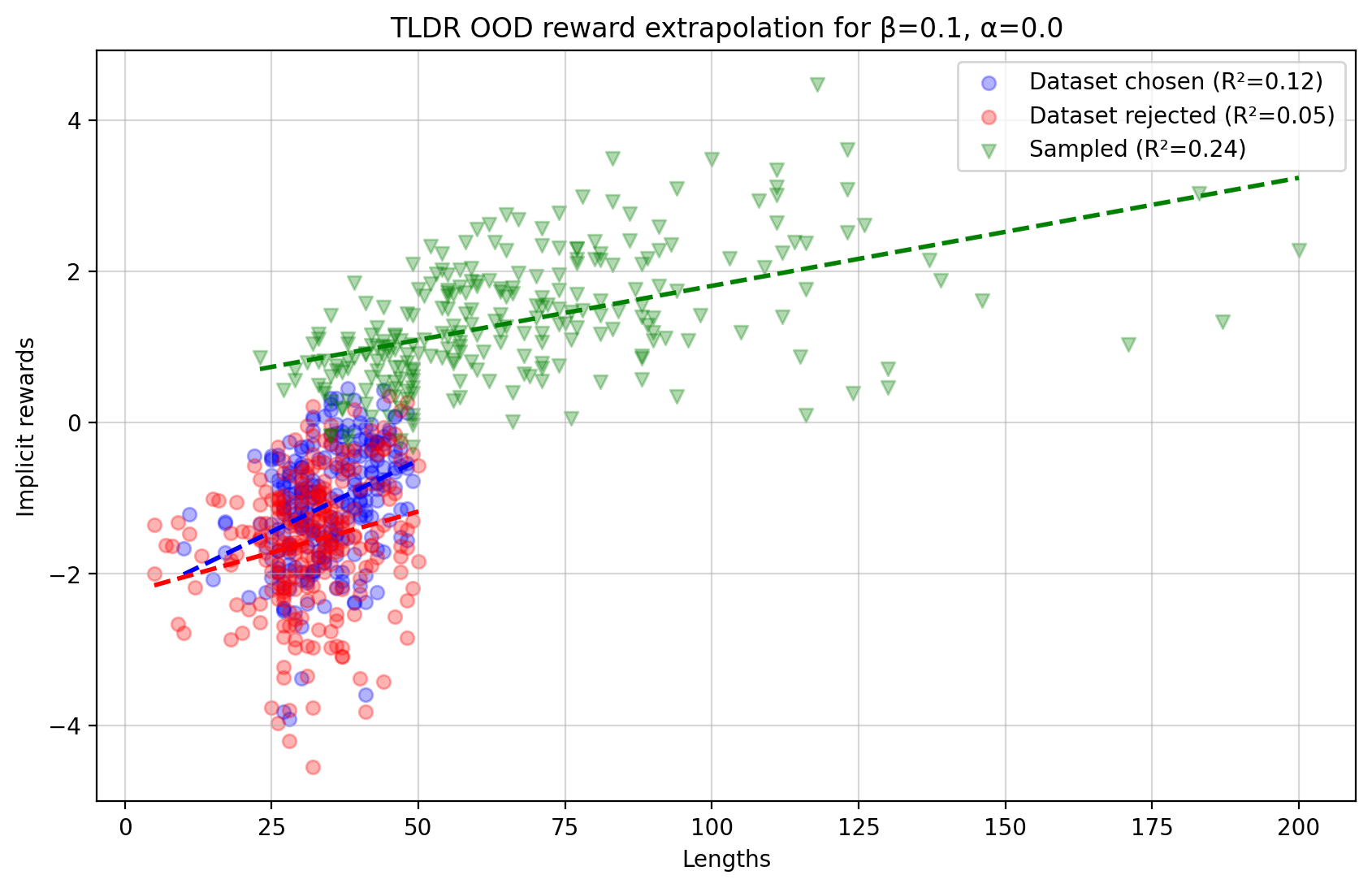}
    \includegraphics[width=0.325\textwidth]{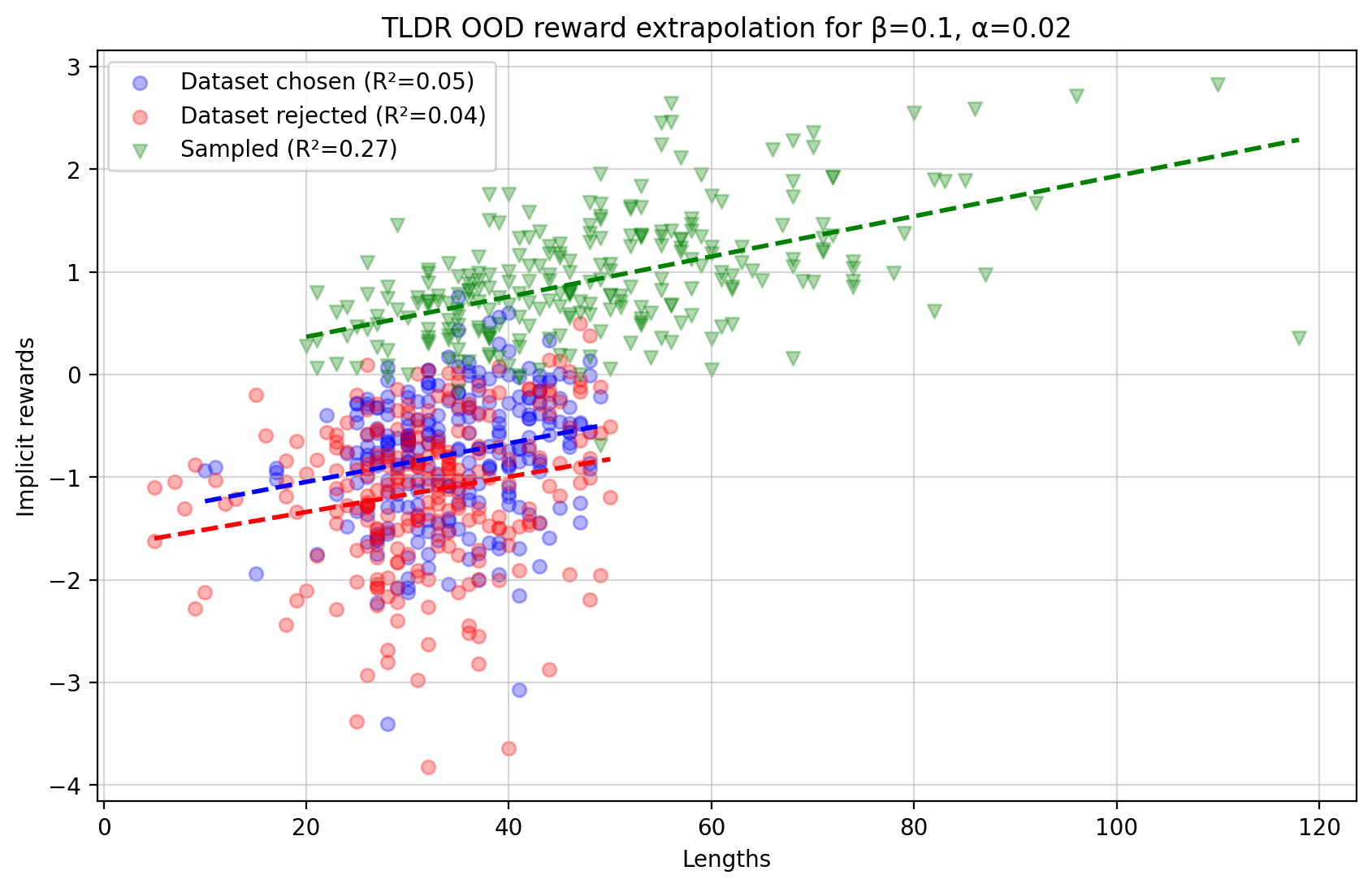}
    \includegraphics[width=0.325\textwidth]{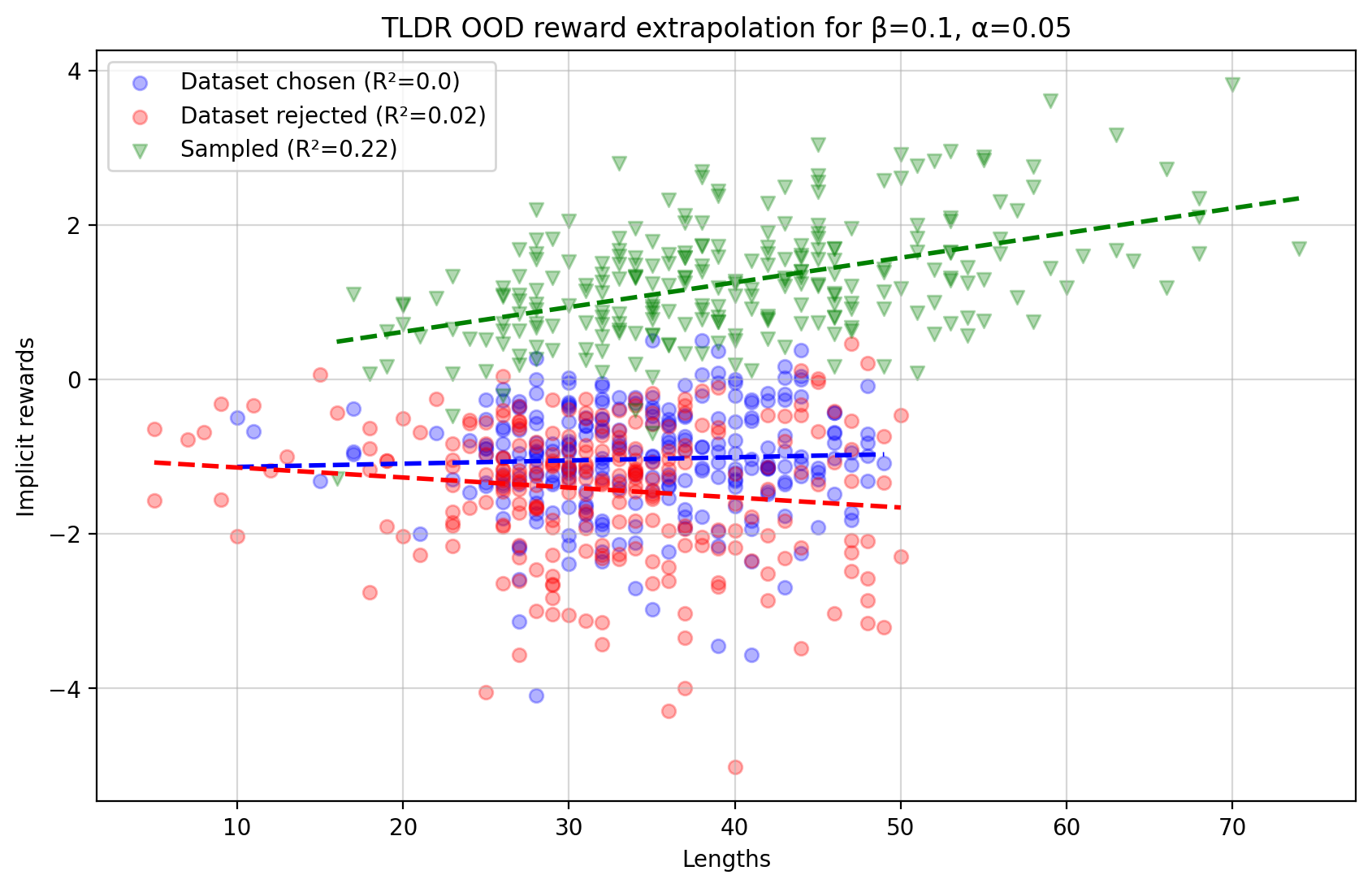}
    \caption{Evaluation of the DPO implicit reward model as defined in Section \ref{sec:DPO} on in-distribution preferred (blue) and rejected (red) answers, as well as OOD answers (green) generated from the corresponding policy. The reward model exhibits little to no length bias in distribution, but significant length correlation outside its training distribution.}
    \label{fig:ood}
\end{figure*}

In classical RLHF, excessive model verbosity \cite{openAI2022} has been well understood as a reward exploitation problem \cite{gao2022scaling, casper2023open, lambert2023alignment}, driven by a bias in the feedback datasets for longer answers. In particular, in the classical RLHF pipeline from Section \ref{sec:rlhf_pipeline}, the reward model is continuously queried on new data generated by the model, creating an out-of-distribution (OOD) robustness issue. These results do not directly transfer to the DPO algorithm, as it does not train a separate reward model and only uses the offline feedback dataset for training. Surprisingly, we find that the exploding length issue in DPO training is similarly driven by out-of-distribution exploitation. We consider the DPO algorithm as an implicit reward training method, as outlined in Section \ref{sec:DPO}. We investigate the behavior of the implicit reward $r_{\theta}$ as defined in Eq. \ref{eq:implicit_reward}. Since the DPO policy $\pi_{\theta}$ is the optimal solution to the constrained RL problem in Eq. \ref{eq:RL} corresponding to $r_{\theta}$, any exploitation behaviour from the policy must be driven by the reward function. In Fig. \ref{fig:ood}, we evaluate $r_{\theta}$ trained with $\beta=0.1$ and different $\alpha$ parameters on the offline feedback dataset (within its training distribution) and on answers generated by the corresponding DPO policy (out of distribution). Surprisingly, within distribution, the corresponding implicit reward models exhibit weak to no length correlation (and even negative length correlation with strong $\alpha$ regularization). However, they all show significant length bias out-of-distribution, with length explaining 30-46\% of the reward variance (as measured by the $R^2$ of a linear regression of the implicit DPO reward on answer length).

\section{Related Work}
In this section we outline relevant work on reward exploitation in RLHF and the verbosity bias.

\noindent\textbf{Reward Exploitation in RLHF:} RLHF reward exploitation, also known as reward over-optimization, is a well-known issue \citep{skalse2022defining, pan2022effects, casper2023open, lambert2023alignment} in which during the reinforcement learning stage, the expected reward keeps improving, but the quality of the model begins to degrade after some point. These effects were confirmed analytically in controlled experiments \cite{gao2022scaling}, as well as empirically in user studies \cite{dubois2024alpacafarm}. Increased model verbosity has been explicitly linked to this phenomenon \cite{openAI2022}. A number approaches have been proposed to mitigate this issue, such as penalizing epistemic uncertainty \cite{coste2023reward, zhai2023uncertaintypenalized, ahmed2024scalable} or using mixture reward models \cite{moskovitz2023confronting}, but they do not explicitly target the length issue.

\noindent\textbf{Mitigating Length Biases in RLHF:} A number of works have sought to explicitly address length biases in RLHF policies. \cite{ramamurthy2023reinforcement} suggest setting a simple discount factor, which improves naturalness of the generated language, \cite{singhal2023long} carry out an extensive study of length correlations in classical RLHF and suggest a number of heuristic-based mitigating approaches. The closes to our approach are the works of \cite{shen2023loose} and the concurrent work of \cite{chen2024odin}, which propose to disentangle length-biases from quality during the reward modeling stage. Our work can be seen as a DPO equivalent counter-part to these approaches. 

As far as we are aware, this is the first work to study the length exploitation problem for Direct Alignment Algorithms such as DPO.

\section{Conclusion}
In this work, we study the problem of length exploitation in the Direct Preference Optimization (DPO) algorithm, extending its analysis from classical RLHF to DPO for the first time. On two standard human feedback datasets, we empirically show that DPO exhibits significant length hacking across a range of hyperparameters. We then specifically link this phenomenon to out-of-distribution bootstrapping. We derive an analytical length-regularized version of the DPO algorithm and show empirically that we can maintain model performance, as evaluated by GPT-4 without significant increases in verbosity, boosting length-corrected win rates by up to 15-20\%. Given the strong length bias in public feedback datasets and the prominence of DPO in the open source community, we hypothesize that many open-source models suffer from similar length-exploitation issues, driving the observations of Fig. \ref{fig:length-rank}. Our results are encouraging, suggesting that open-source models could match proprietary ones on automated evaluations on a length corrected basis as well.

\newpage
\section{Limitations}
Our work addresses the particular issue of length exploitation in Direct Preference Optimization. Our regularization objective requires explicit penalty function (such as length) and may not be suitable to avoid general exploitation issues along axes separate from verbosity. Furthermore, we only study the DPO objective, which might behave differently from other direct alignment algorithms, which use different objective functions. We also e4valuate our approach on one model size and two smaller-scale public datasets of human feedback. It is unclear what the scaling laws of such exploitation behaviours might be and to what degree they are dependent on model size, capability and data quality beyond length biases. We believe these questions are a promising direction for future work.

\section*{Ethics Statement}
This work focuses on alleviating and empirical extrapolation issues during DPO training, specifically an increasing verbosity bias. Experiments are ran on publicly-available data and pre-trained models, and do not release any new models for public use. As a result, there should be no ethical concerns.

\section*{Acknowledgements}
Chelsea Finn is a CIFAR Fellow in the Learning in Machines and Brains program. This work was also supported by ONR grant N00014-22-1-2621 and the Volkswagen Group.

\bibliography{custom}
\bibliographystyle{acl_natbib}

\appendix
\begin{table*}[t]
\centering
\begin{tabular}{lccccccc}
    \toprule
    \textbf{Dataset} &
      \multicolumn{3}{c}{\textbf{MT Bench Score}} &
      \multicolumn{3}{c}{\textbf{Sample Length}}  \\
      & {Turn 1} & {Turn 2} & {Mean} & {Turn 1} & {Turn 2} & {Mean} \\
    \cmidrule(r){2-4}\cmidrule(l){5-7}
    Regularized DPO ($\alpha=0.05$) & \textbf{7.29} & 5.71 & \textbf{6.50} & 295.15 & 243.26 & 269.21 \\
Standard DPO ($\alpha=0.0$) & 7.14 & \textbf{5.81} & 6.48 & 292.00 & 260.03 & 276.01 \\
SFT & 6.84 & 5.00 & 5.92 & \textbf{257.88} & \textbf{246.69} & \textbf{252.28} \\

    \bottomrule
\end{tabular}
\caption{Phi-2 Ultrafeedback results on $\beta = 0.1$ and $\alpha \in [0, 0.05]$. Bold indicates best value (highest score, lowest length). All evaluations done with standard MT bench parameters, see \url{https://github.com/lm-sys/FastChat/blob/main/fastchat/llm_judge/README.md} for more details.}
\label{table:phi2}
\end{table*}

\section{Phi-2 UltraFeedback Experiments}

To further validate the effectiveness of the proposed length regularization schena, we run a small set of experiments with the 2.7B Microsoft model Phi-2 \cite{javaheripi2023phi} on the Ultrafeedback binarized dataset\footnote{\url{https://huggingface.co/datasets/HuggingFaceH4/ultrafeedback_binarized}} \cite{cui2023ultrafeedback}. This dataset consists of 64K prompts, whose completions are generated by LLMs and then ranked with GPT. The chosen response is considered to be the highest-scoring completion, and the rejected prompt is chosen at random from the other 3 responses. We use the same experimental setup as for the TL;DR/HH Pythia 2.8B experiments (see Appendix \ref{sec:exp_details}), and run 1 epoch of supervised fine-tuning on Ultrafeedback prior to alignment with DPO. We compare $\alpha=0.0$ and $\alpha=0.05$, using $\beta=0.1$ for both models. For evaluation, we use the multi-turn MT Bench harness \cite{zheng2023judging}. The results (Table \ref{table:phi2}) indicate that the length regularization strategy decreases length, while actually increasing downstream performance, though both gains are small.

\section{Additional Out-of-Distribution Experiments}

\begin{figure*}
    \centering
    \includegraphics[width=0.325\textwidth]{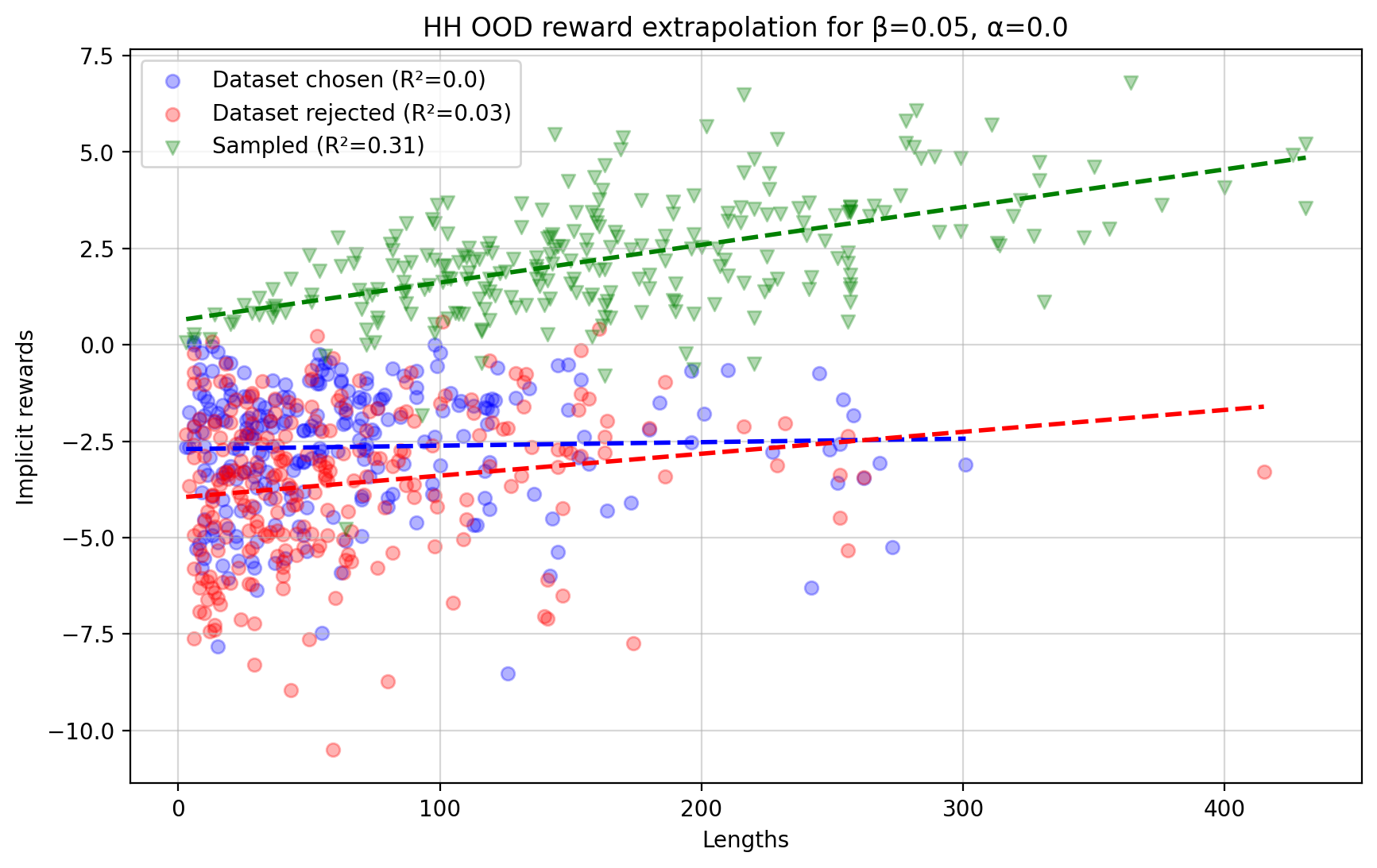}
    \includegraphics[width=0.325\textwidth]{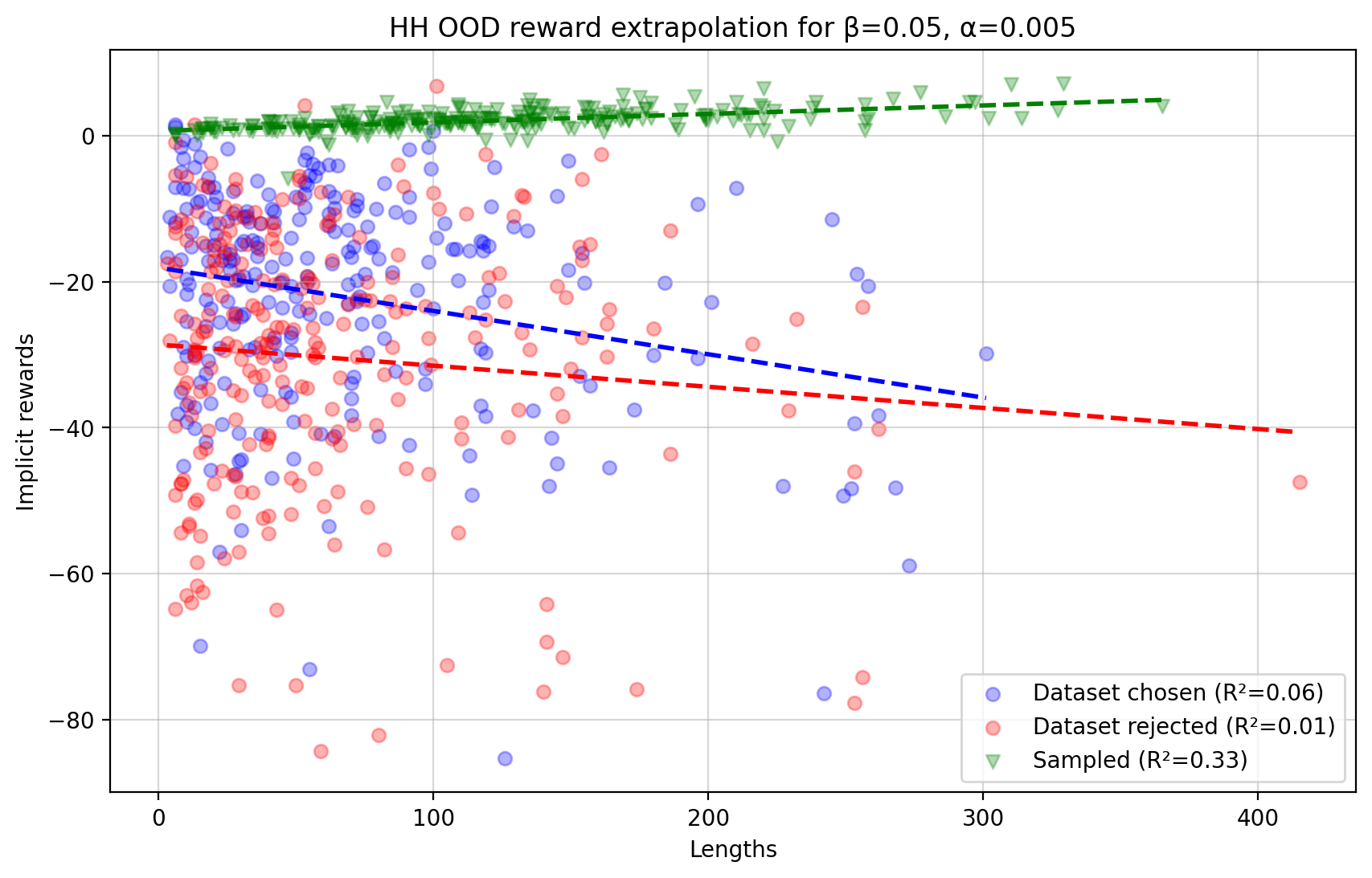}
    \includegraphics[width=0.325\textwidth]{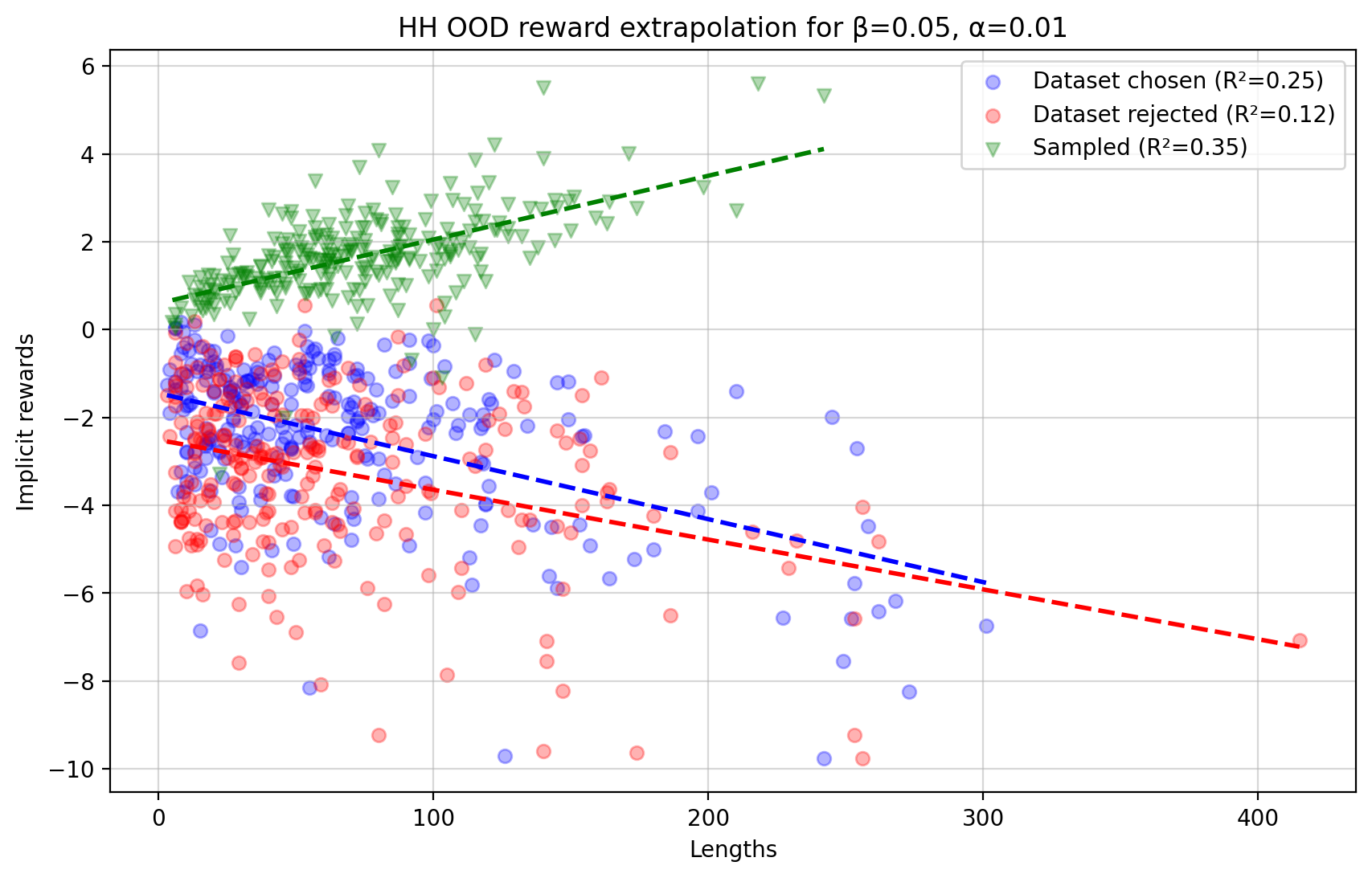}
    \includegraphics[width=0.325\textwidth]{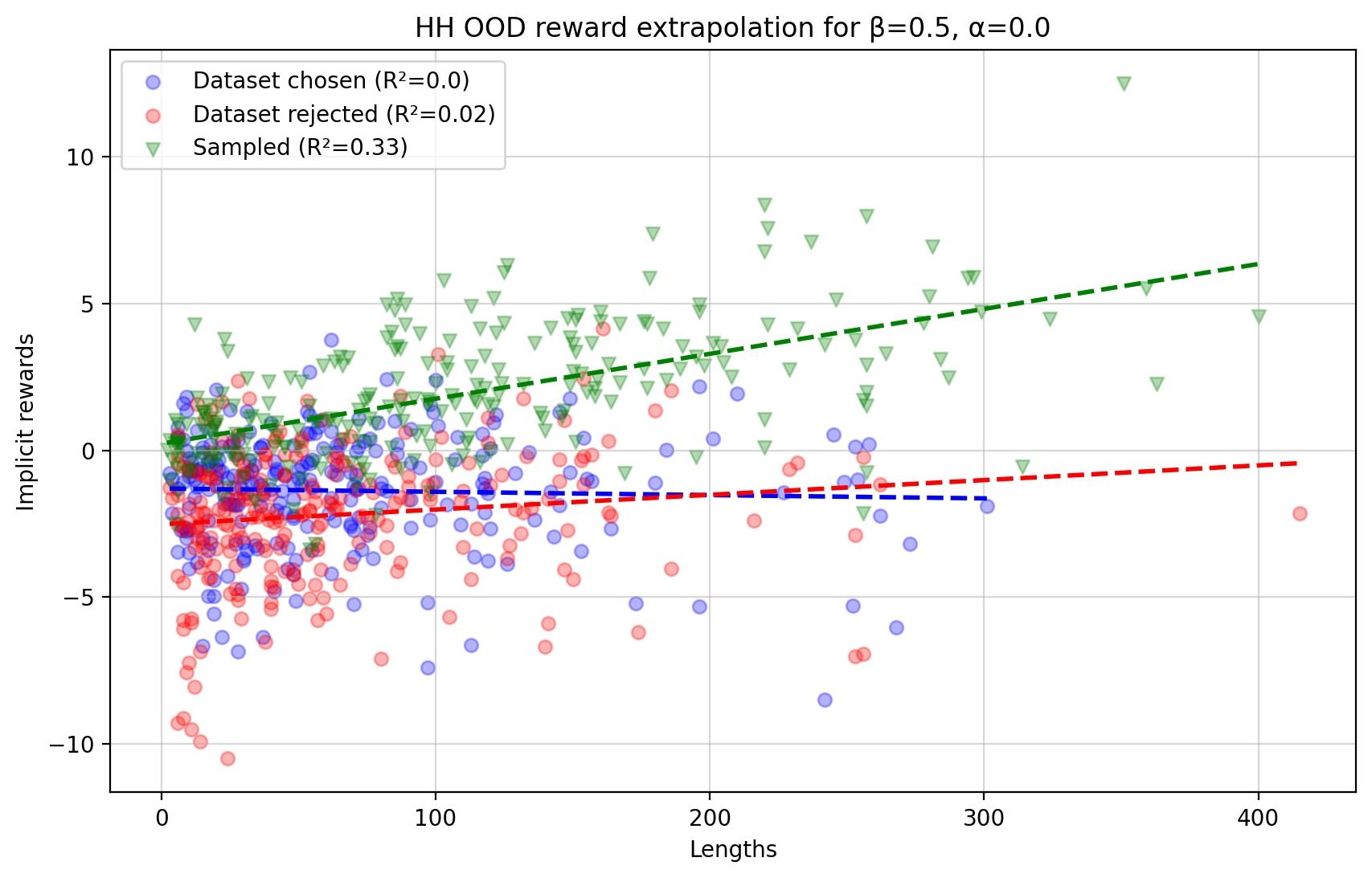}
    \includegraphics[width=0.325\textwidth]{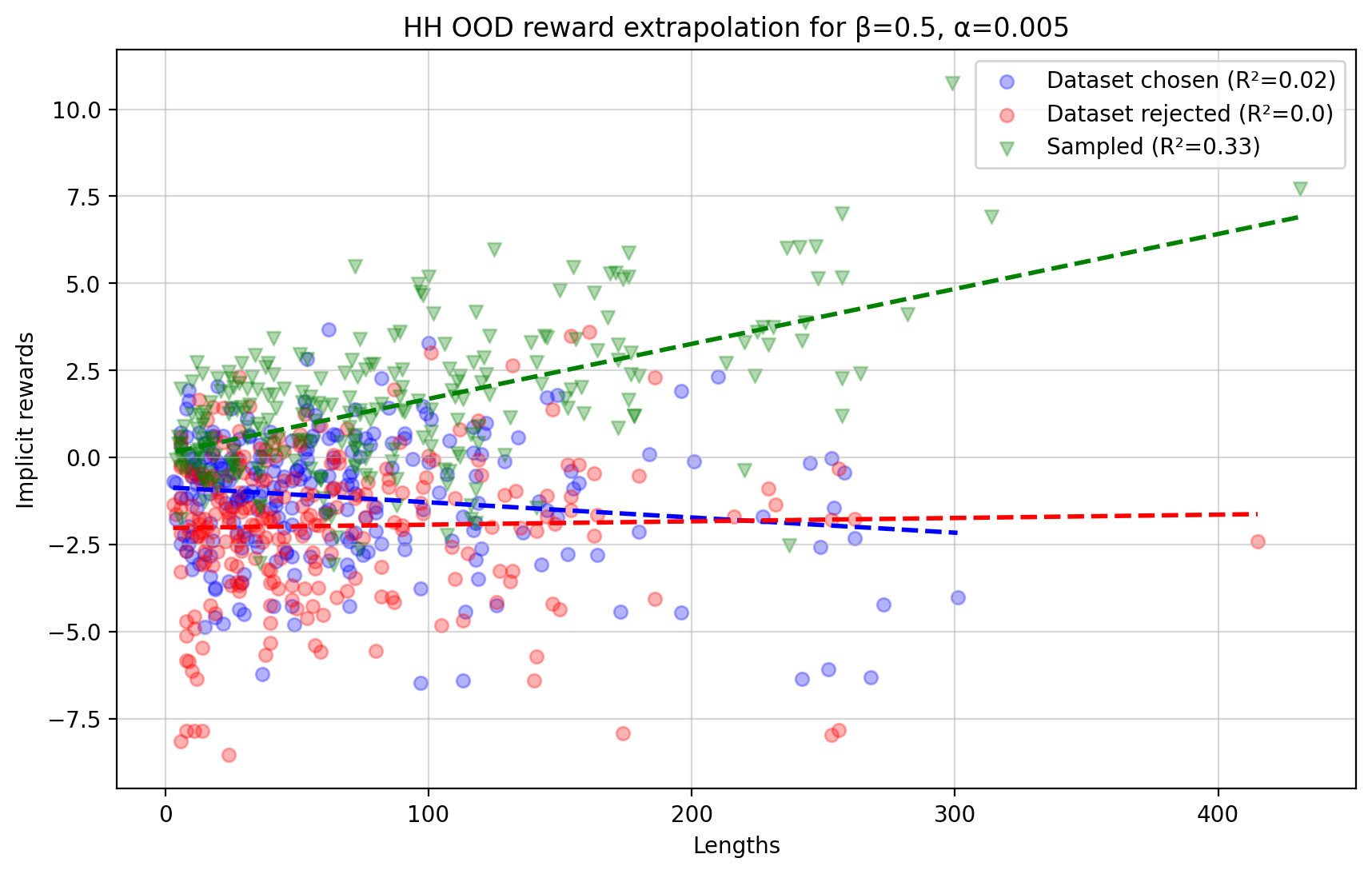}
    \includegraphics[width=0.325\textwidth]{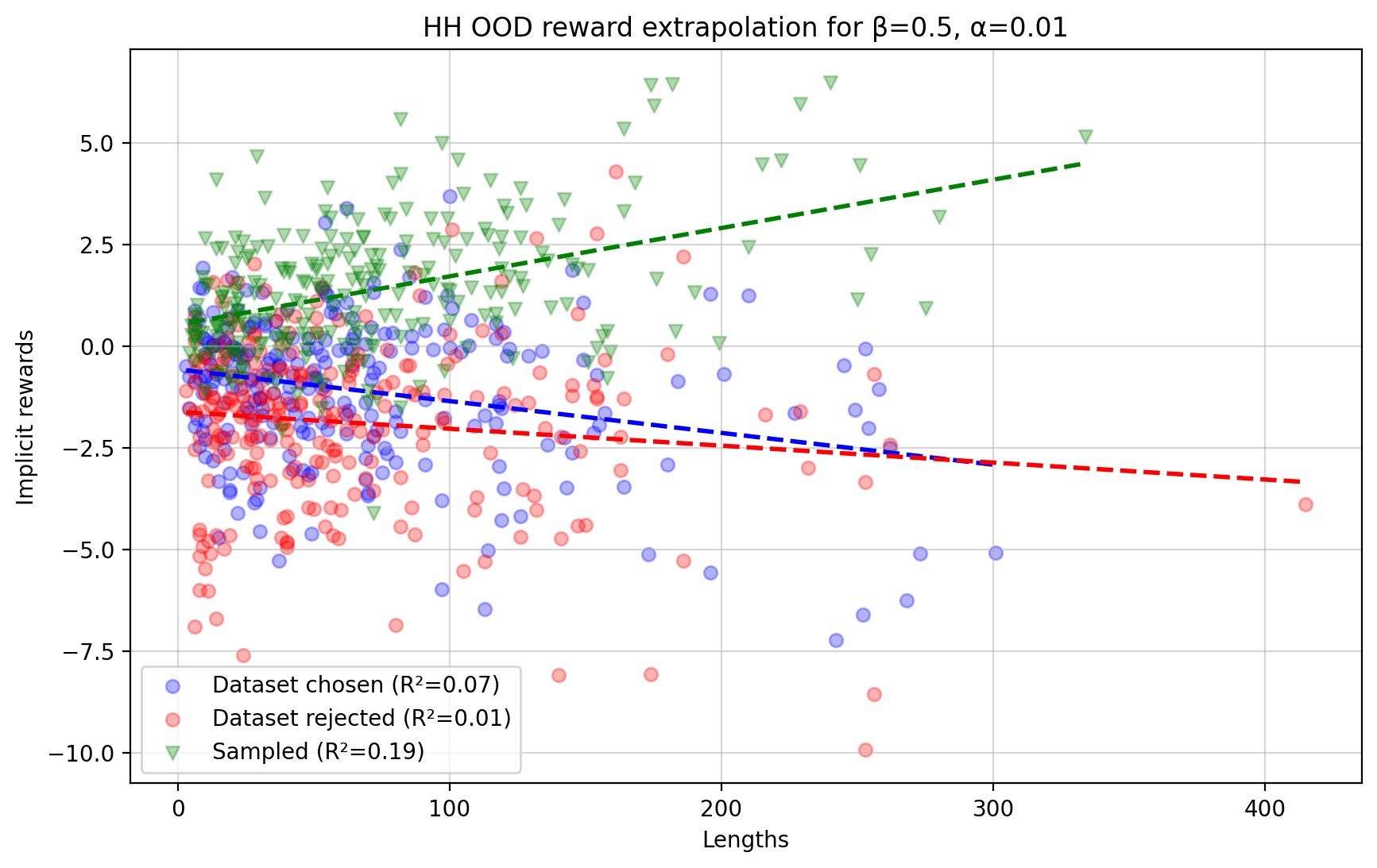}
    \caption{HH KL divergence and DPO implicit reward (Eq. \ref{eq:implicit_reward}) evaluated on dataset responses (preferred in blue, dispreferred in red) as well as model-generated responses (green). \textbf{Top row:} $\alpha=0.0$, \textbf{Middle row:} $\alpha=0.005$, \textbf{Bottom row:} $\alpha=0.01$. \textbf{Left column:} $\beta=0.05$, \textbf{Right column:} $\beta=0.5$.}
    \label{fig:ood_appendix_hh}
% \end{figure*}
% \begin{figure*}
    \vspace{20mm}
    \centering
    \includegraphics[width=0.325\textwidth]{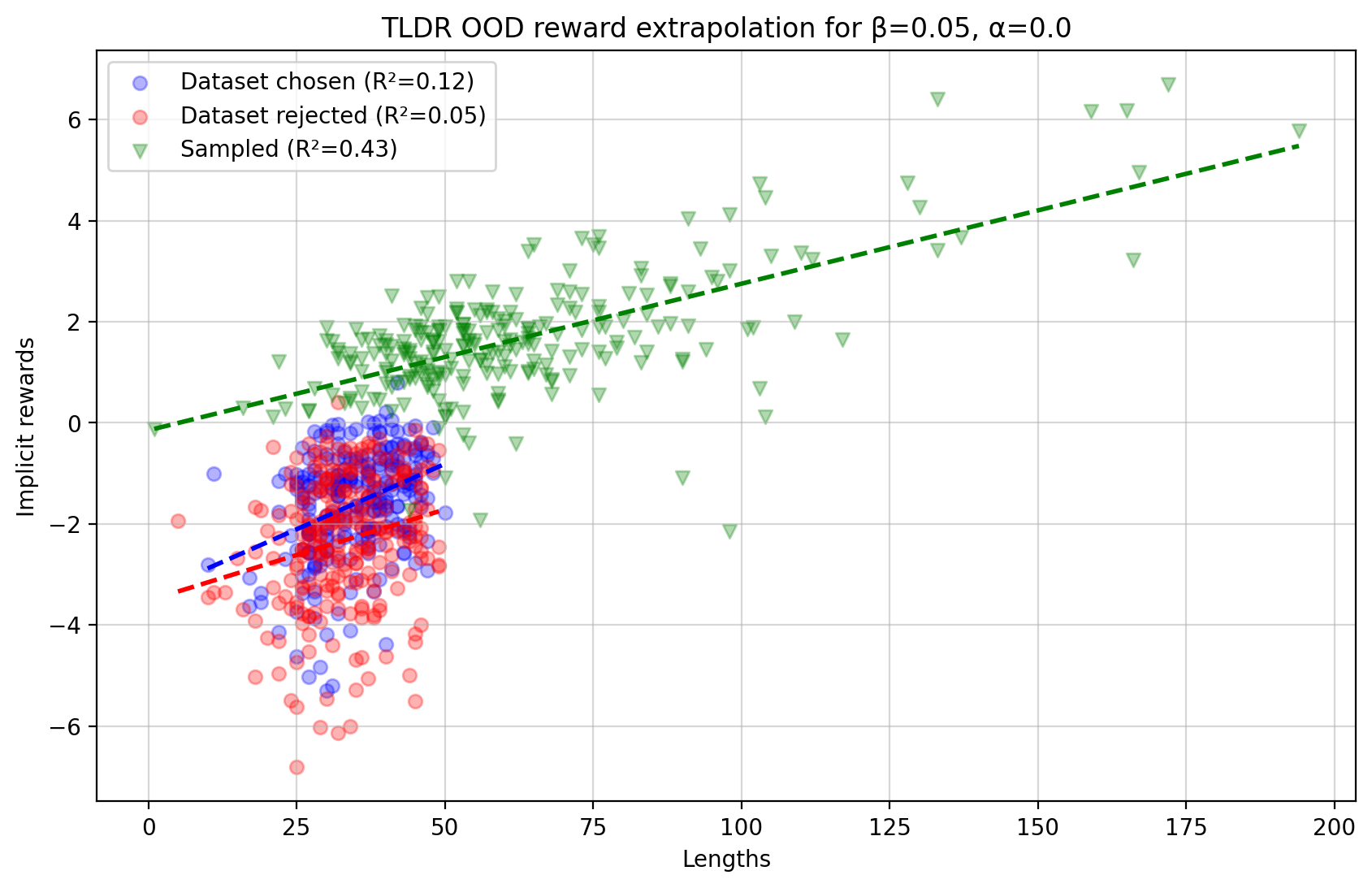}
    \includegraphics[width=0.325\textwidth]{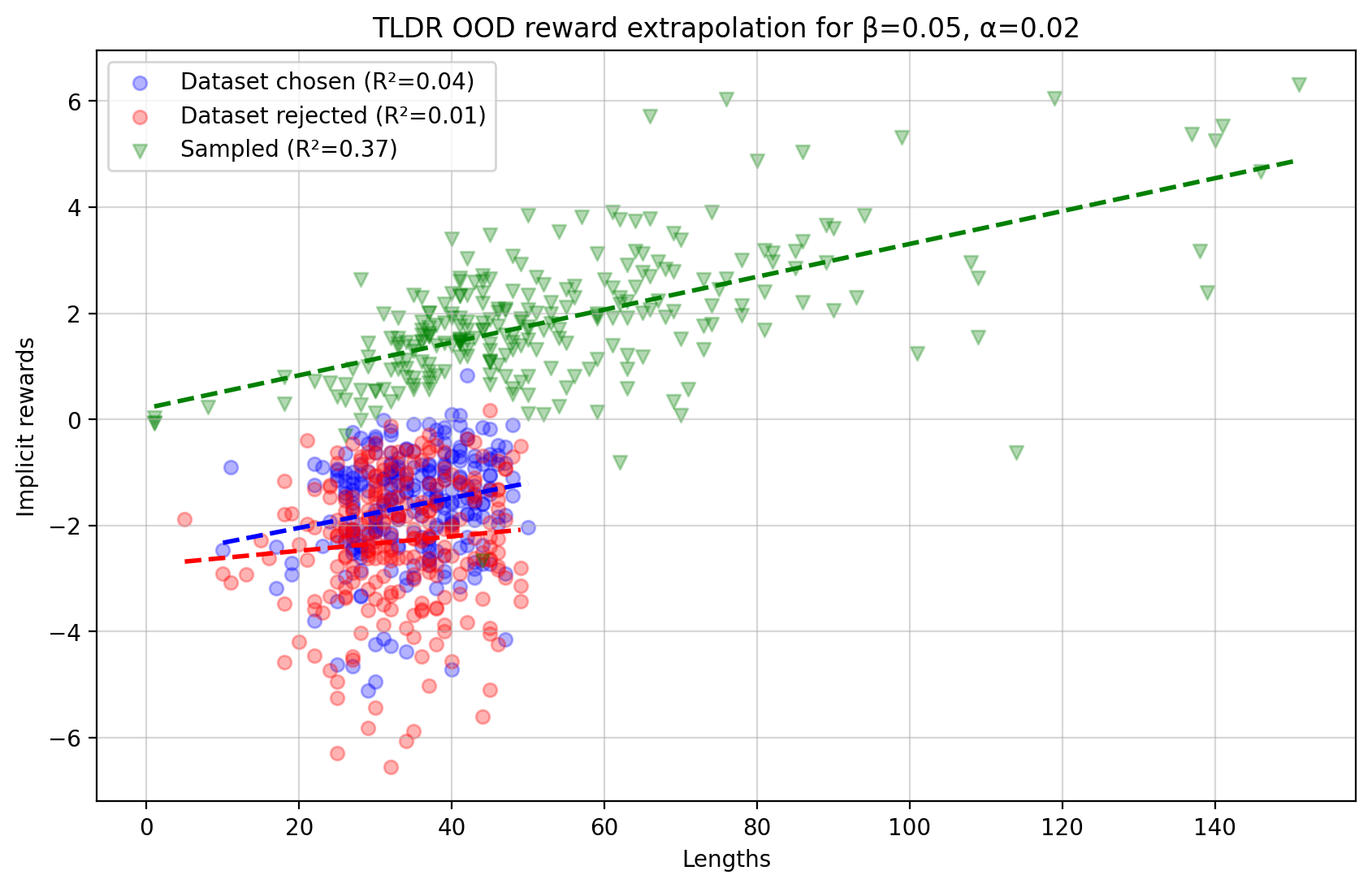}
    \includegraphics[width=0.325\textwidth]{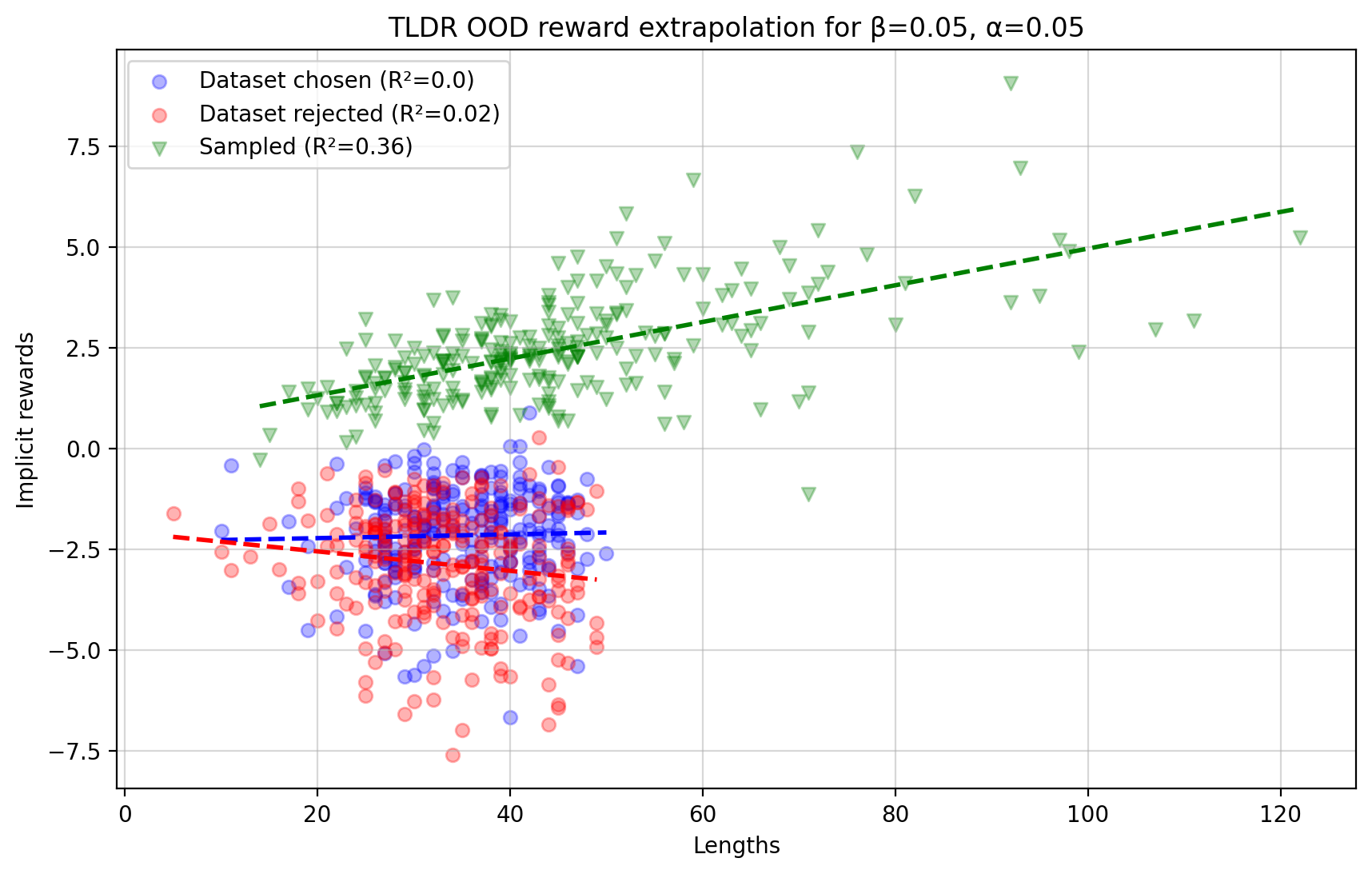}
    \includegraphics[width=0.325\textwidth]{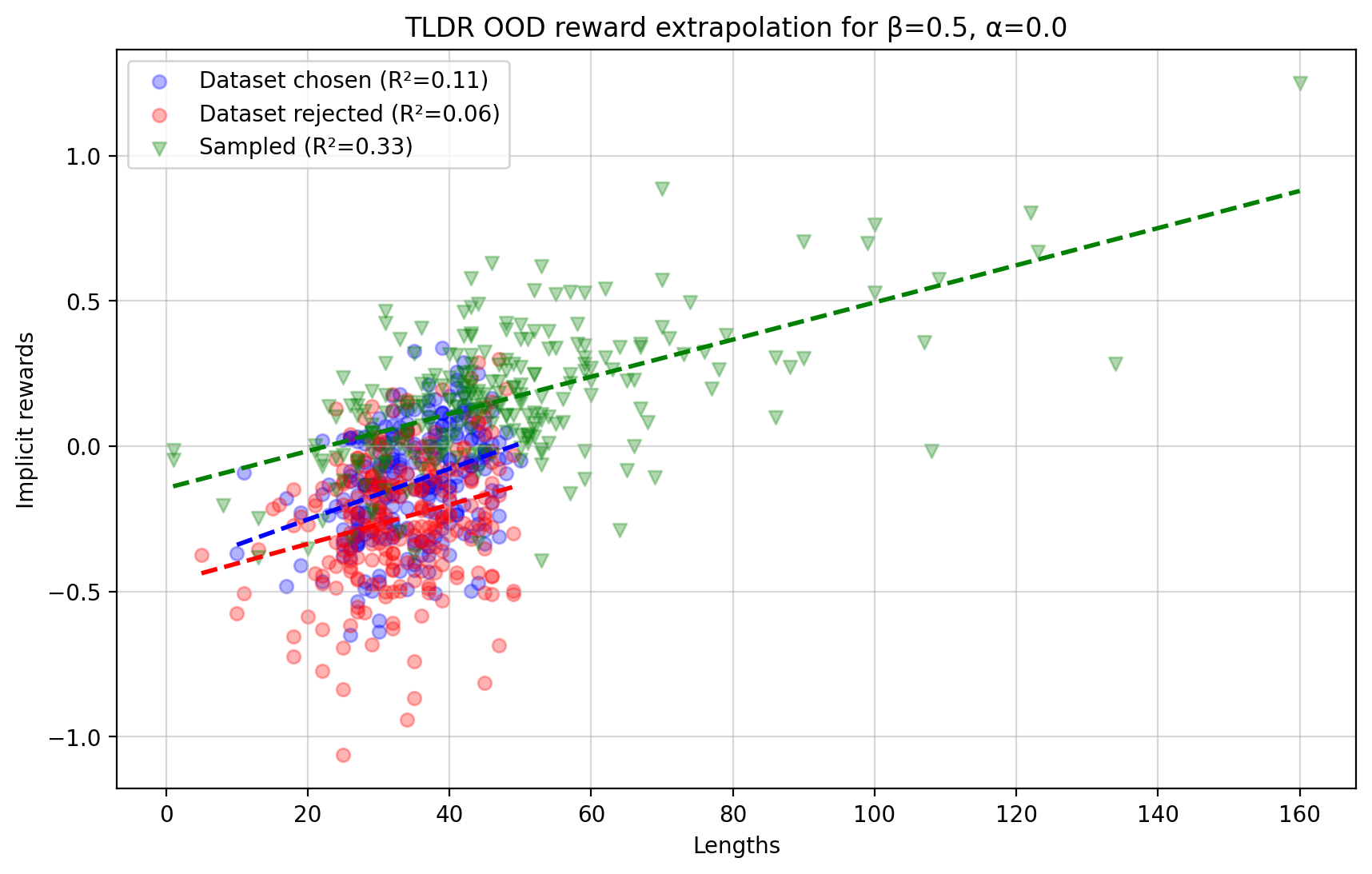}
    \includegraphics[width=0.325\textwidth]{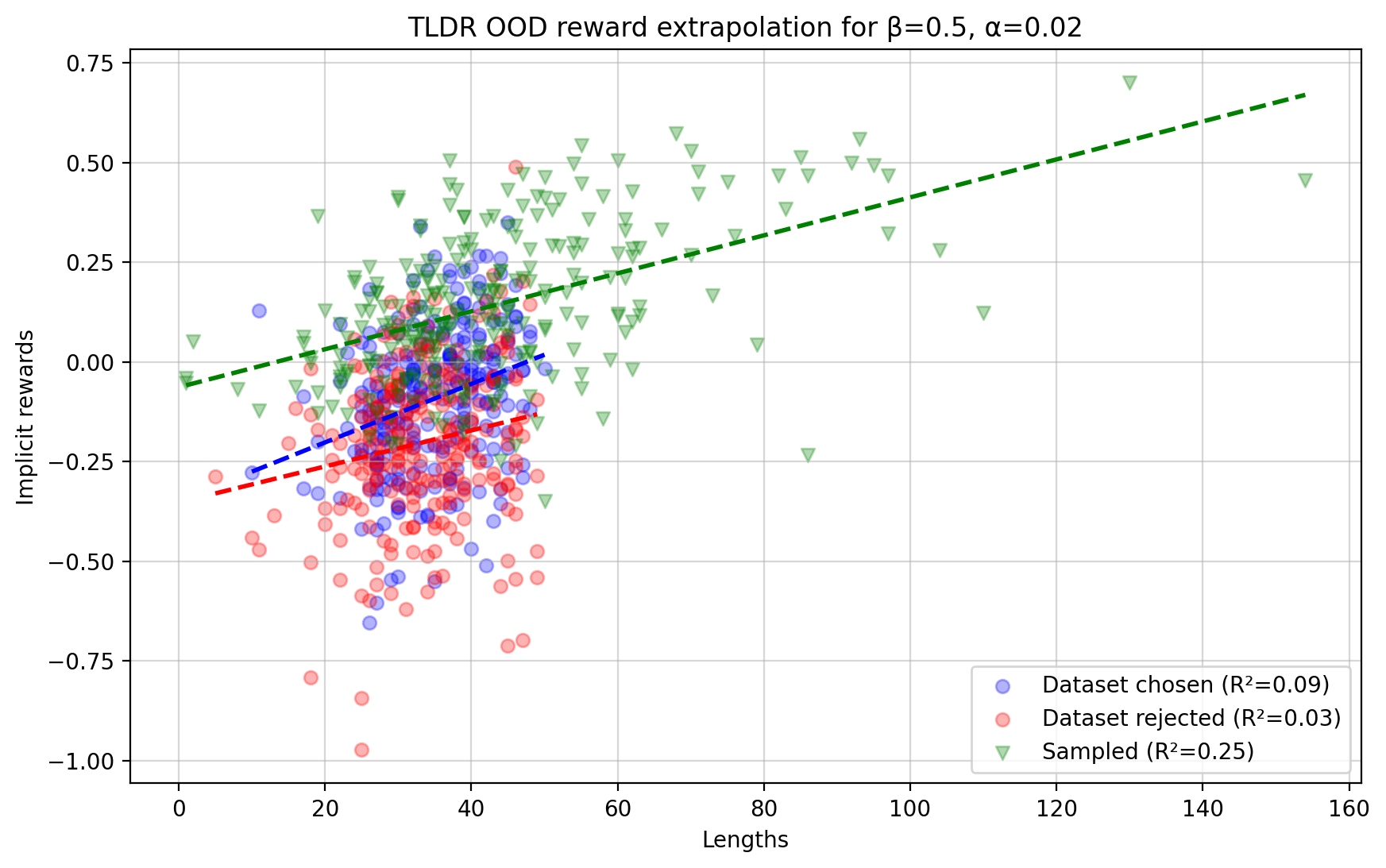}
    \includegraphics[width=0.325\textwidth]{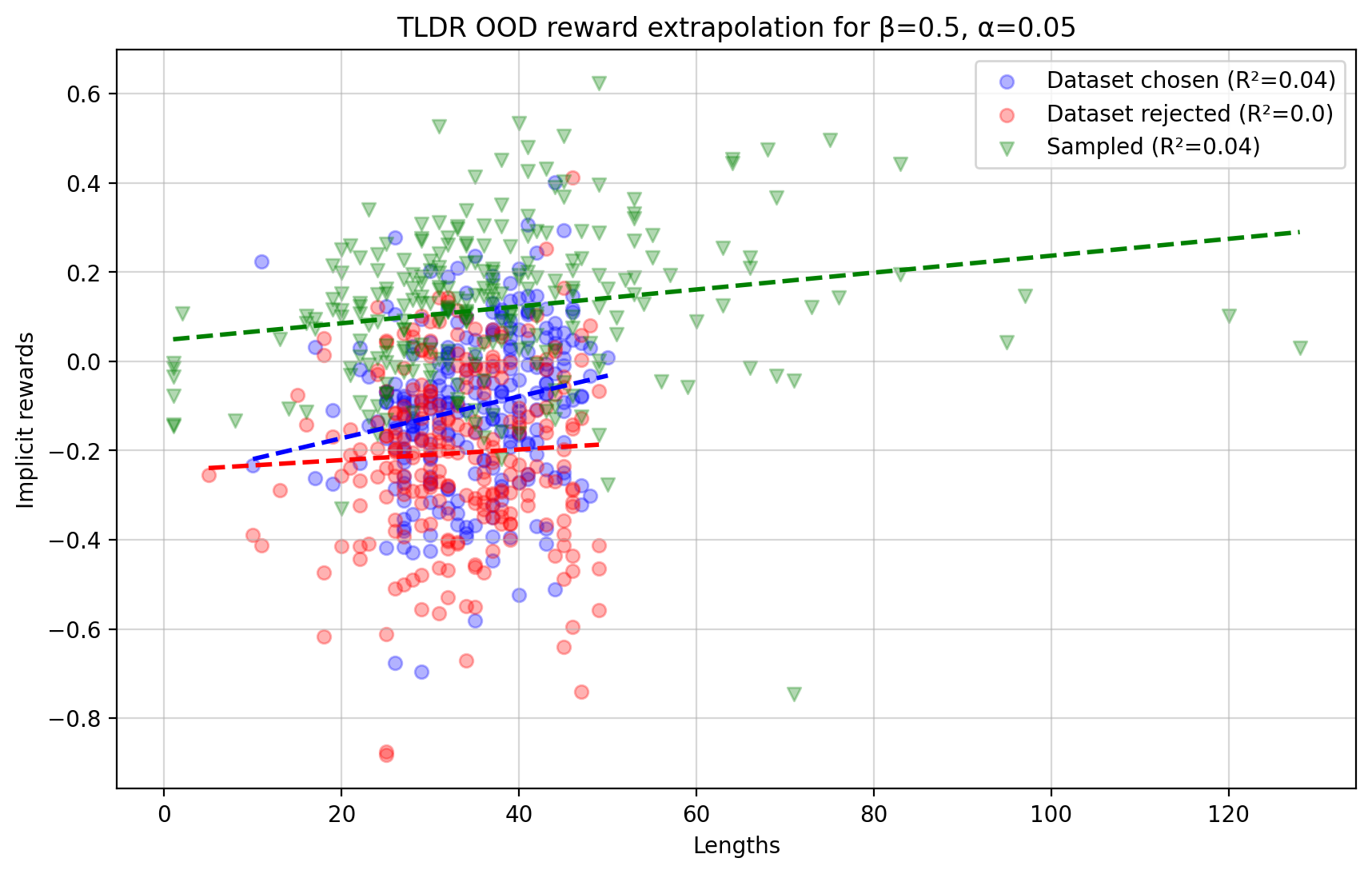}
    \caption{TL;DR KL divergence and DPO implicit reward (Eq. \ref{eq:implicit_reward}) evaluated on dataset responses (preferred in blue, dispreferred in red) as well as model-generated responses (green). \textbf{Top row:} $\alpha=0.0$, \textbf{Middle row:} $\alpha=0.02$, \textbf{Bottom row:} $\alpha=0.05$. \textbf{Left column:} $\beta=0.05$, \textbf{Right column:} $\beta=0.5$.}
    \label{fig:ood_appendix_tldr}
\end{figure*}
In Fig. \ref{fig:ood_appendix_hh} and Fig. \ref{fig:ood_appendix_tldr}, we provide the rest of the reward-length correlation experiments across both TL;DR and HH (see Fig. \ref{fig:ood}). The results across different $\alpha$ and $\beta$ parameter values indicate a similar pattern of out-of-distribution extrapolation along the length axis. 
\section{Experimental Details}
\label{sec:exp_details}

We follow the original DPO codebase (\href{https://github.com/eric-mitchell/direct-preference-optimization}{https://github.com/eric-mitchell/direct-preference-optimization}) with default hyperparameters unless otherwise noted (1 epoch of training, batch size of 128 with 16 gradient accumulation steps, RMSProp with a learning rate of $0.5 \times 10^{-6}$ with linear warm-up for 150 steps).

For the experiments in the main body of the paper, we used the OpenAI TL;DR dataset (92K preferred/dispreferred response pairs) \cite{stiennon2022learning}. Each prompt is a Reddit post belonging to one of several topic forums, with title/post metadata included. 256 prompts sampled from the held-out set are used for all evaluations (e.g. loss, accuracy, KL, winrates, length), with temperature $1.0$ and max length $512$. All models in the main experiments were initialized from Pythia 2.8B base pre-trained weights, and underwent supervised fine-tuning on TL;DR for 1 epoch prior to DPO. All experiments were carried out on 4 NVIDIA A40 GPUs for a total of about 2000 GPU hours. All evaluations were computed with \texttt{gpt-4-1106-preview} as judge, with random positional flips to avoid known bias. We use the same GPT-4 evaluation prompts as in \cite{rafailov2023direct}, they are listed below for completeness.
\\[3mm]
\textbf{HH GPT-4 winrate prompt.}
\begin{verbatim}
For the following query to a chatbot, which 
response is more helpful?

Query: <the user query>

Response A:
<either the test method or baseline>

Response B:
<the other response>

FIRST provide a one-sentence comparison of the 
two responses and explain which you feel is 
more helpful. SECOND, on a new line, state only 
"A" or "B" to indicate which response is more 
helpful. Your response should use the format: 
Comparison:
<one-sentence comparison and explanation>
More helpful: <"A" or "B">
\end{verbatim}
\noindent\textbf{TL;DR GPT-4 winrate prompt.}
\begin{verbatim}
Which of the following summaries does a better 
job of summarizing the most important points 
in the given forum post?

Post:
<post>

Summary A:
<Summary A>

Summary B:
<Summary B>

FIRST provide a one-sentence comparison of 
the two summaries, explaining which you prefer 
and why.  SECOND, on a new line, state only
"A" or "B" to indicate your choice.
Your response should use the format:
Comparison: 
<one-sentence comparison and explanation>
Preferred: <"A" or "B">
\end{verbatim}

\section{Sample Responses Across Models}

In this section, we provide examples of model responses generated by length-regularized DPO (\texttt{RDPO}), standard DPO (\texttt{DPO}), and the initial policy (\texttt{SFT}), in response to prompts from the HH/TL;DR evaluation sets. These examples are selected to highlight cases where regularized DPO wins against the dataset response (according to the GPT-4 evaluator), but both the DPO and SFT-produced answers do not. All samples were generated with the standard procedure in Appendix \ref{sec:exp_details} from the same set of 256 prompts used for the rest of evaluations. See Tables \ref{table:1} through \ref{table:4} for samples.

\begin{table*}[ht]
    \centering
    \begin{tabular}{p{0.2\linewidth} | p{0.75\linewidth}}
      \toprule
      \textbf{Prompt} & Human: How often should a one year old shower?

Assistant: An important part of daily hygiene is to shower or bathe on a regular basis. The frequency of bathing or showering depends on the weather and activity. Children should bathe or shower as needed to keep clean. Generally children under one year of age don’t need to bathe or shower. When children are one year old, you should bathe or shower as often as your child’s doctor recommends. Generally, it’s fine to take your child for a bath or shower as often as you feel like.

Human: Okay, but too often might strip the oils from their skin.

Assistant:
        \\
      \midrule
      \textbf{SFT} & Correct, taking the wrong age appropriate number of baths or showers can cause the oil layer to thin or disappear.  Generally speaking, it’s best to bathe or shower your child for a healthy clean skin.  You might need to bathe or shower your child after using soap or other cleansing products.\\
      \midrule
      \textbf{DPO} & You’re correct! Ideally, a child should bathe or shower only when they need to, but there are some occasions when parents may want to take a more frequent approach to washing their child. For example, a child under one year old might benefit from being bathed or showered more frequently if their skin is oily or if they have a common skin irritant or infection. As a general rule, it’s best to take a more frequent approach to bathing or showering your child when they are young, rather than a more infrequent one. You should still only bathe or shower as often as your child’s doctor recommends. For your child’s health and wellbeing, you should be as careful and responsible about bathing or showering as possible.\\
      \midrule
      \textbf{RDPO} & You’re right, taking a child for a bath too often can strip the child’s skin of its oils. Bathing too frequently can also cause skin irritation and rashes, so try to get your child into the shower or bath only as needed. \\
    \end{tabular}
    \caption{HH prompt, $\beta=0.05$, $\alpha\in[0, 0.05]$.}
    \label{table:1}
\end{table*}

\begin{table*}[ht]
    \centering
    \begin{tabular}{p{0.2\linewidth} | p{0.75\linewidth}}
      \toprule
      \textbf{Prompt} & Human: Should you rent a Uhaul to move?

Assistant: Do you need to transport very large, heavy items?

Human: Yes like couches and beds.

Assistant: I would probably recommend a Uhaul.

Human: Can anyone drive one of them?

Assistant:
        \\
      \midrule
      \textbf{SFT} & No. \\
      \midrule
      \textbf{DPO} & Well yes and no.  It’s not easy, but you might be able to rent one from your gas station or auto parts store, or you could buy the driver’s manual for one from Amazon for around \$60, and you can always learn if you don’t feel confident enough.  The main challenge is that you’ll need to be very sure that you’re fully competent before driving one.  You might also consider renting one, because it’s also better for the environment because it’s more efficient.  So for example, if you need to transport two couches instead of one, it will take you two trips instead of one.  And there’s less plastic packaging in the world, so that’s good for the environment. \\
      \midrule
      \textbf{RDPO} & Renting a Uhaul from an authorized rental company requires a driver’s license and a minimum of 2.5 years of driving experience.  So no, it’s not possible to rent one without having a driver’s license and 2.5 years of driving experience. \\
    \end{tabular}
    \caption{HH prompt, $\beta=0.1$, $\alpha\in[0, 0.05]$.}
    \label{table:2}
\end{table*}

\begin{table*}[ht]
    \centering
    \begin{tabular}{p{0.2\linewidth} | p{0.75\linewidth}}
      \toprule
      \textbf{Prompt} & SUBREDDIT: r/relationships
      
      TITLE: Me[25M] with my EX GF [18F] 2 years, We broke up because I flirted with another girl.
      
      POST:
\# Part 1

I'm finding it hard to understand something. My ex GF broke up with me 6 months ago, I tried winning her back for 6 months. We would see each other once every 2 weeks and we'd get on like 2 peas in a pod and she kept leading me on, however 2 days after seeing her she would tell me it was a mistake and that we can't speak anymore. This of course was really unfair and caused me great depression. 

\# Part 2

After dealing with this for 6 months I decided to completely cut her out, and I've been doing good by not messaging her anymore. She still messages me almost every 2nd night. I don't get it.. She doesn't want anything to do with me however she still messages me. Surely if someone wanted to cut someone out of their life they wouldn't go through all the trouble of constantly messaging someone? 

\# Part 3

I guess what I'm trying to get at is, does anyone have an explanation as to why someone who broke up with me keeps messaging me even after I said goodbye for good? 

Thanks\\
      \midrule
      \textbf{SFT} &  \\
      \midrule
      \textbf{DPO} & in advance 

Me (24f) and ex GF of 6 months (29f) broke up 6 months ago. She still messes me up when I tell her we can never speak again and I've been deleting all her text messages and emails. Why does she still message me?
 \\
      \midrule
      \textbf{RDPO} & SO confused on why GF keeps messaging after saying she doesn't want anything to do with me beyond the odd text. \\
    \end{tabular}
    \caption{TL;DR prompt, $\beta=0.05$, $\alpha\in[0, 0.05]$.}
    \label{table:3}
\end{table*}

\begin{table*}[ht]
    \centering
    \begin{tabular}{p{0.2\linewidth} | p{0.75\linewidth}}
      \toprule
      \textbf{Prompt} & SUBREDDIT: r/relationships
      
      TITLE: Me [40 F] with my husband [40 M] . . . we have been married 17 years. He has absolutely no interest in going out on date nights, not even for our anniversary or for my birthday.
      
      POST: I'm just hurting inside. Number one, he doesn't want to spend time with me without the kids. Number two, it doesn't matter to him at all that I would like some time without the kids. You'd think that even if he didn't want to go out, he'd do it at least once a year for me, you know?

We moved here, 400 miles away from my family, so he could take his dream job. Between work and school he is often gone 80 hours a week. One of our children is special needs with sensory processing disorder and anxiety/depression. I do most of the parenting while trying to run a freelance business on the side. 

My family would often help with the kids but without the family support and with him being gone it's so hard. And he won't compromise. He just doesn't ever want to go out on date nights. He'll happily go to a baseball game for work or go out golfing with his buddies, but he'll never do anything for me. And I get really sad that he doesn't seem to crave any alone time with me.  

Am I being unreasonable to want to get out on a date night? What should I do?
\\
      \midrule
      \textbf{SFT} & my husband takes forever to get to date night and never wants to spend time with me. \\
      \midrule
      \textbf{DPO} & My boyfriend of two years isn't interested in me spending time on a date night with him. Instead, I've found him working long hours with his dream job and I am alone all the time. I feel like I am dying inside because we used to go out regularly and now he doesn't seem to even want that type of relationship. I just feel unloved by this man and I need some answers. BF of two years has chosen to pursue some crazy dream job instead of spending time with me on a regular basis. I am lonely and unhappy. I need some clarity on what I can do about this issue. \\
      \midrule
      \textbf{RDPO} & My husband seems to take very little time out for me and I feel neglected. He rarely devotes time for date night and I feel deprived and lonely. I want him to be more invested in our relationship. \\
    \end{tabular}
    \caption{TL;DR prompt, $\beta=0.1$, $\alpha\in[0, 0.05]$.}
    \label{table:4}
\end{table*}

\end{document}